\documentclass[10pt,journal,letterpaper,compsoc]{IEEEtran}

\usepackage{times}
\usepackage{graphicx}
\usepackage{amsmath}
\usepackage{amssymb}
\usepackage{esvect}
\usepackage{bm}
\usepackage{bbm}
\usepackage{enumerate}
\usepackage[lined,boxruled,linesnumbered]{algorithm2e}
\usepackage{epstopdf}
\usepackage{multirow}
\usepackage{subcaption}

\begin{document}
%
\title{\huge Modeling and Inferring Human Intents and Latent Functional Objects for Trajectory Prediction}
\author{Dan~Xie, Tianmin~Shu,~Sinisa~Todorovic and~Song-Chun~Zhu 
\IEEEcompsocitemizethanks{
\IEEEcompsocthanksitem D. Xie and T. Shu are with Department of Statistics, University of California, Los Angeles. Email: \{xiedan,tianmin.shu\}@ucla.edu.
\IEEEcompsocthanksitem S. Todorovic is with School of EECS, Oregon State University. Email: sinisa@eecs.oregonstate.edu.
\IEEEcompsocthanksitem S.-C. Zhu is with Department of Statistics and Computer Science, University of California, Los Angeles. Email: sczhu@stat.ucla.edu.}
\thanks{}}

\markboth{For review: IEEE transactions on pattern analysis and machine intelligence}%
{Shell \MakeLowercase{\textit{et al.}}: Modeling and Inferring Human Intents and Latent Functional Objects for Trajectory Prediction}
\IEEEcompsoctitleabstractindextext{%
\begin{abstract}

This paper is about detecting functional objects and inferring human intentions in surveillance videos of public spaces. People in the videos are expected to intentionally take shortest paths toward functional objects subject to obstacles, where people can satisfy certain needs (e.g., a vending machine can quench thirst), by following one of three possible intent behaviors: reach a single functional object and stop, or sequentially visit several functional objects, or initially start moving toward one goal but then change the intent to move toward another.  Since detecting functional objects in low-resolution surveillance videos  is typically unreliable, we call them ``dark matter'' characterized by the functionality to attract people. We formulate the Agent-based Lagrangian Mechanics wherein human trajectories are probabilistically modeled as motions of agents in many layers of ``dark-energy'' fields, where each agent can select a particular force field to affect its motions, and thus define the minimum-energy Dijkstra path toward the corresponding source ``dark matter''. For evaluation, we compiled and annotated a new dataset. The results demonstrate our effectiveness in predicting human intent behaviors and trajectories, and localizing functional objects, as well as discovering distinct functional classes of objects by clustering human motion behavior in the vicinity of functional objects.

\end{abstract}
\begin{keywords}
scene understanding, video analysis, functional objects, intents modeling, human motion
\end{keywords}}

\maketitle

\IEEEdisplaynotcompsoctitleabstractindextext
\IEEEpeerreviewmaketitle

\section{Introduction}\label{sec:intro}
\subsection{Motivation and Objective}\label{motivation}

\begin{figure}
\centering
\includegraphics[width=0.9\linewidth]{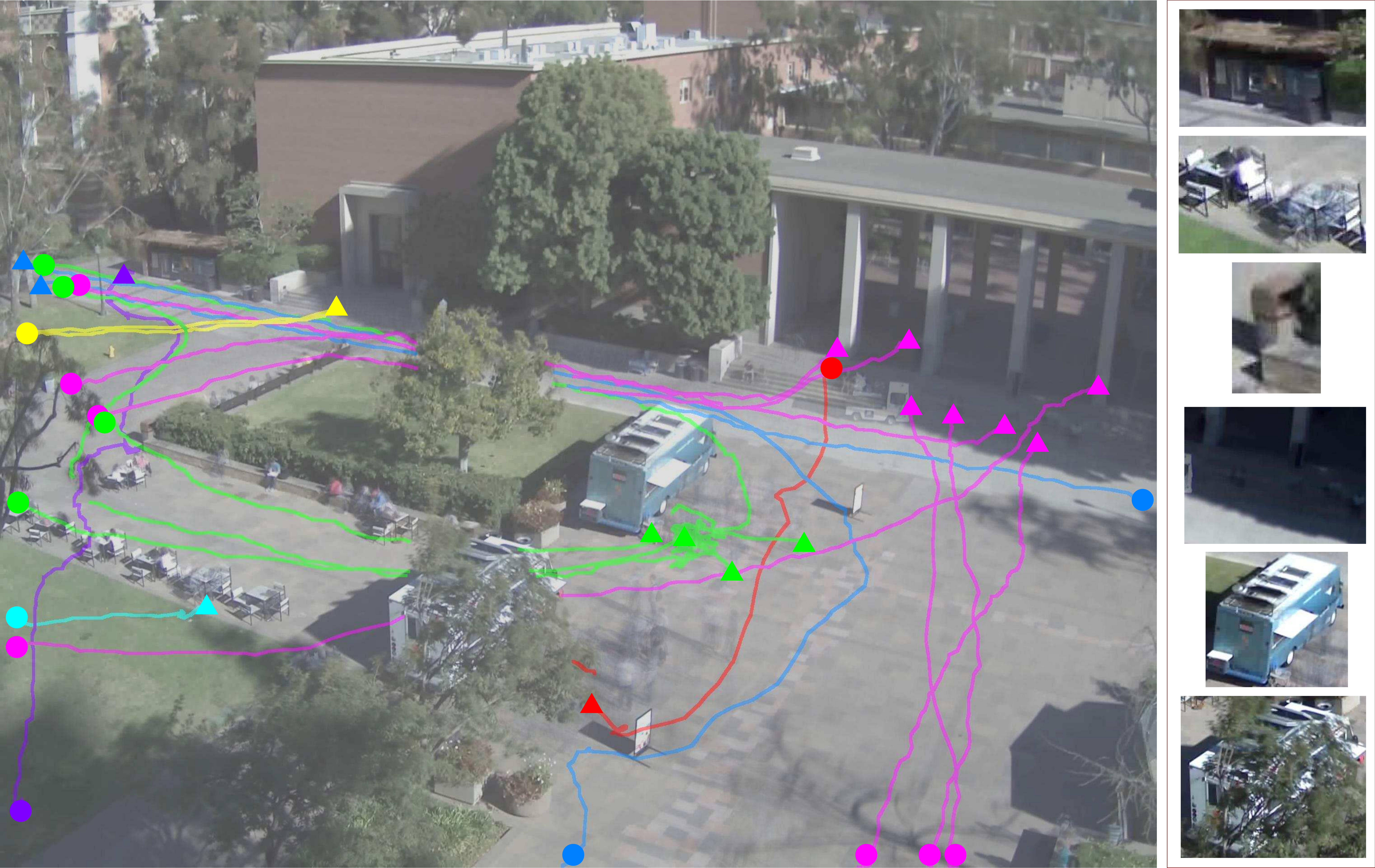}
\caption{({\em left}) People's trajectories are color-coded by their shared goal destination. The triangles denote destinations, and the dots denote start positions of the trajectories. E.g., people may be heading toward the food-truck to buy food (green), or the vending machine to quench thirst (blue). ({\em right}) Due to low resolution, poor lighting, and occlusions, objects at the destinations are very difficult to detect only based on their appearance and shape.}
\label{fig:example_trajectory}
\end{figure}

\IEEEPARstart{T}his paper presents an approach that seeks to infer why and how people move in surveillance videos of public spaces. Regarding the ``why'', we expect that people typically have certain needs (e.g., to quench thirst, satiate hunger, get some rest), and hence intentionally move in open spaces toward certain destinations where their needs can be satisfied (e.g., vending machine, food truck, bench), as illustrated in Fig.~\ref{fig:example_trajectory}. These goal destinations are occupied by objects whose functions in the scene are defined in relation to the corresponding human needs. Tab. \ref{tab:field_need} lists some additional examples of human needs and functional objects that can satisfy them considered in this paper. Regarding the ``how'', we make the assumption that people take shortest paths to intended destinations, while avoiding obstacles and non-walkable surfaces. We also consider three types of human intent behavior, including: ``single intent'' when a person reaches a single functional object and stops, ``sequential intent''  when a person wants to sequentially visit several functional objects (e.g., buy food at the food-truck, and go to a bench to have lunch), and ``change of intent'' when a person initially starts moving toward one goal but then changes the intent to move toward another (e.g. because the line in front of the food-truck is too long).

The answers to the above ``why'' and ``how'' people move in a video can be used to predict entire human trajectories in the future after observing only their initial short parts. Also, detecting functional objects and reasoning about human trajectories in their vicinity can be used for discovering distinct functional classes of objects in public spaces.

Thus, given videos of open spaces, our problem involves:
\begin{itemize}
\item Prediction of
\begin{itemize}
\item Latent human intents and intent behaviors,
\item Latent functional map of the scene, i.e., locations of functional objects and non-walkable surfaces,
\item How partially observed human trajectories fully unfold in the unobserved future,
\end{itemize}
\item Discovery of functional classes of objects.
\end{itemize}

\begin{table}
\begin{center}
\begin{tabular}{|c|c|}
\hline
Functional object & Human need \\
\hline\hline
Vending machine / Food truck / Table & Hunger \\ \hline
Water fountain / Vending machine  & Thirst\\ \hline
ATM / Bank & Money\\ \hline
Chair / Table / Bench /  Grass & Rest\\ \hline
News stand / Ad billboard & Information\\ \hline
Trash can & Hygiene\\ \hline
Bush / Tree & Shade from the sun \\ \hline
\end{tabular}
\end{center}
\caption{Examples of human needs and objects in the scene that can satisfy these needs in the context of a public space. Each object may attract a person to approach it, and repel another person to stay away from it.}
\label{tab:field_need}
\end{table}


While in some scenes functional objects could be identified based on their appearance, and consequently their functionality inferred from the recognized semantic classes, our domain presents two fundamental challenges to such an approach, which motivates us to formulate an alternative, novel framework.

First,  functionality classes are not tightly correlated with semantic object classes, because instances of the same object may have different functionality  in our scenes (e.g., a bench may attract people to get some rest, or repel them if it is freshly painted). Also, the same object may be the goal destination to some people and a deterrent to others.

Second, the low-resolution of our surveillance videos does not allow for reliable object detection, even by a human eye (see Fig.~\ref{fig:example_trajectory}). Hence, locations of functional objects in the scene appear as ``dark matter'' emanating ``dark energy'' which affects people. We use this terminology to draw an analogy to cosmology, where dark matter is hypothesized to account for a large part of the total mass in the universe. Analogous to our poorly visible functional objects exerting attraction and repulsion forces on people, dark matter is invisible, and its existence and properties are inferred from its gravitational effects on visible matter.


From above, the key contribution of this paper involves a joint representation and inference of:
\begin{itemize}
\item Visible domain--- traditional recognition categories: objects, scenes, actions and events;  and
\item Functional domain --- higher level cognition concepts: fluent, causality, intents, attractions and physics.
\end{itemize}

To formulate this problem, we leverage the framework of Lagrange mechanics, and introduce the concept of {\em field}, in the same way as in physics, such as the gravitational, electric and magnetic fields. Each functional object and non-walkable surface in the scene generates an attraction (positive) and repulsion (negative) field. Thus, we view the scene as a physical system populated by particle-agents who move in many layers of ``dark-energy'' fields. Unlike inanimate particles, each agent can select a particular force field to affect its motions, and thus define the minimum-energy Dijkstra path toward the corresponding source ``dark matter''. In the following, we introduce the main steps of our approach.

\begin{figure*}
\centering
\includegraphics[width=0.75\linewidth]{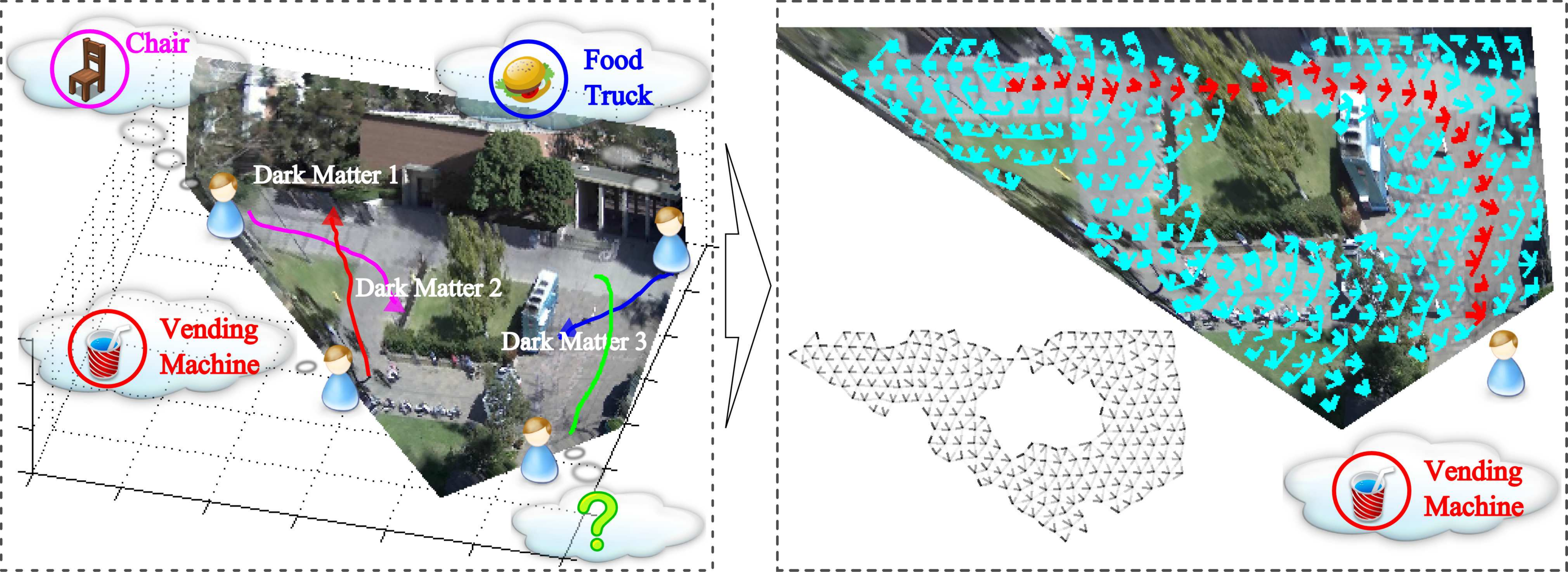}
\caption{An example video where people driven by latent needs move toward functional objects  where these needs can be satisfied (i.e., ``dark matter''). (Right) A zoomed-in top-down view of the scene and our actual results of: (a) Inferring and localizing the person's goal destination; (b) Predicting the person's full trajectory (red); (c) Estimating the force field affecting the person (the blue arrows, where their thickness indicates the force magnitude; the black arrows represent another visualization of the same field.); and (d) Estimating the constraint map of non-walkable areas and obstacles in the scene (the ``holes'' in the field of blue arrows and the field of black arrows).}
\label{fig:example_scene}
\end{figure*}

\subsection{Overview of Our Approach} \label{sec:overview}


%
Fig.~\ref{fig:example_scene} illustrates main steps of our approach.

{\bf Tracking.} Given a short video excerpt, we first extract people's trajectories using the state-of-the-art multitarget tracker of \cite{Pirsiavash2011} and the low-level  3D scene reconstruction of \cite{Zhao2011}.  While the tracker and 3D scene reconstruction perform well, they may yield noisy results. Also, these results represent only partial observations, since the tracks of most people in the given video are not fully observable, but get cut out at the end. These noisy, partial observations are used as input features to our inference.

{\bf Bayesian framework.} Uncertainty is handled by specifying a joint pdf of observations, latent layout of non-walkable surfaces and functional objects, and people's intents and trajectories. Our model is based on the following  assumptions. People are expected to have only one goal destination at a time, and be familiar with the scene layout (e.g., from previous experience), such that they can optimize their trajectory as a shortest path toward the intended functional object, subject to the constraint map of non-walkable surfaces. We consider three types of intent behavior. A person may change the intent and decide to switch to another goal destination, have only a single intent, or want to sequentially reach several functional objects.

{\bf Agent-based Lagrangian Mechanics.} Our Bayesian framework leverages the Lagrangian mechanics (LM) by treating the scene as a physics system where people can be viewed as charged particles moving along the mixture of repulsion and attraction energy fields generated by obstacles and functional objects. The classical LM, however, is not directly applicable to our domain, because it deterministically applies the principle of Least Action, and thus provides a poor model of human behavior.

We extend LM to an agent-based Lagrangian mechanics (ALM) which accounts for latent human intentions. Specifically, in ALM, people can be viewed as charged particle-agents with capability to intentionally select one of the latent fields, which in turn guides their motions by the principle of Least Action.

{\bf Inference.} We use the data-driven Markov Chain Monte Carlo (MCMC) for inference \cite{Tu2002, Kwon2013}. In each iteration, the MCMC probabilistically samples the number and locations of obstacles and sources of ``dark energy'', and people's intents. This, in turn, uniquely identifies the ``dark energy'' fields in the scene. Each person's trajectory is estimated as the globally optimal Dijkstra path in these fields, subject to obstacle constraints. The predicted trajectories are used to estimate if they arose from ``single'', ``sequential'' or ``change'' of human intents. In this paper, we consider two inference settings: {\em offline} and {\em online}. The former first infers the layout of ``dark matter'' and obstacles in the scene as well as people's intents, and fixes these estimates for predicting Dijkstra trajectories and human intent behavior. The latter, on the other hand, sequentially estimates both people's intents and human trajectories frame by frame, where the estimation for frame $t$ uses all previous predictions.

We present experimental evaluation on challenging, real-world videos from the VIRAT \cite{Oh2011}, UCLA Courtyard \cite{Amer2012}, UCLA Aerial Event \cite{Shu2015} datasets, as well as on our five new videos of public squares. Our ground truth annotations and the new dataset will be made public. The results demonstrate our effectiveness in predicting human intent behaviors and trajectories, and localizing functional objects, as well as discovering distinct functional classes of objects by clustering human motion behavior in the vicinity of functional objects. Since localizing functional objects in videos is a new problem, we compare with existing approaches only in terms of predicting human trajectories. The results show that we outperform prior work on VIRAT and UCLA Courtyard datasets.

\subsection{Relationship to Prior Work}\label{sec:related_work}
This section reviews three related research streams in the literature, including the work on functionality recognition, human tracking, and prediction of events. For each stream, we also point out our differences and contributions.

\textbf{Functionality recognition}. 
Recent work has demonstrated that performance in object and human activity recognition can be improved by reasoning about functionality of objects. Functionality is typically defined as an object's capability to satisfy certain human needs, which in turn triggers corresponding human behavior. E.g., reasoning about how people handle and manipulate small objects can improve accuracy of recognizing calculators or cellphones \cite{Gall2011, Gupta2009}. Some other object classes can be directly recognized by estimating how people interact with the objects \cite{Delaitre2012}, rather than using common appearance features. This interaction can be between a person's hands and the object \cite{Pieropan2013}, or between a human skeleton and the objects \cite{Wei_2013_ICCV}. Another example is the approach that successfully recognizes chairs among candidate objects observed in the image by using human-body poses as context for identifying whether the candidates have functionality ``sittable'' \cite{Grabner2011}. Similarly, video analysis can be improved by detecting and localizing functional scene elements, such as parking spaces, based on low-level appearance and local motion features \cite{Turek2010}. The functionality of moving objects  \cite{Oh2010} and urban road environments \cite{Qin2014} has been considered for advancing activity recognition. 

As in the above approaches,  we also resort to reasoning about functionality of objects based on human behavior and interactions with the objects, rather than use standard appearance-based features. Our key difference is that we {\em explicitly} model latent human intents which can modify an object's functionality -- specifically, in our domain, an object may simultaneously attract some people {\em and} repel others, depending on their intents. 


\textbf{Human tracking and planning}. 
A survey of vision-based trajectory learning and analysis for surveillance is presented in \cite{Morris2008}. The related approaches differ from ours in the following aspects. Estimations of: (a) Representative human motion patterns in (years') long video footage \cite{Abrams2012}, (b) Lagrangian particle dynamics of crowd flows  \cite{Ali2007, Ali2008}, and (c) Optical-flow based dynamics of crowd behaviors \cite{Solmaz2012} do not account for individual human intents. Reconstruction of an unobserved trajectory segment has been addressed only as finding the shortest path between the observed start and end points \cite{Gong2011}. Early work also estimated a numeric potential field for robot path planning  \cite{Barraquand1992}, but did not account for the agents free will to choose and change goal destinations along their paths. Optimal path search \cite{Shao2005}, and reinforcement learning and inverse reinforcement learning \cite{Baker2009a, Ziebart2009, Kitani2012} was used for explicitly reasoning about people's goals for predicting human trajectories. However, these approaches considered: i) Relatively sanitized settings with scenes that did not have many and large obstacles (e.g., parking lots); and ii) Limited set of locations for people's goals (e.g., along the boundary of the video frames). People's trajectories have also been estimated based on inferring social interactions  \cite{Helbing1995, Mehran2009, Pellegrini2009, Alahi2016}, or detecting objects in egocentric videos \cite{Park2016}. However, these approaches critically depend on domain knowledge. For example, the approaches of \cite{Kitani2012} and   \cite{Park2016} use appearance-based object detectors, learned on training data, for predicting trajectories. In contrast, we are not in a position to apply appearance-based object detectors for identifying hidden functional objects, due to the low resolution of our videos. Finally, a Mixture of Kalman Filters has been used to cluster {\em smooth} human trajectories based on their dynamics and start and end points \cite{Zhou2012a}. Instead of linear dynamics, we use the principle of Least Action, and formulate a globally optimal planning of the trajectories. This allows us to handle sudden turns and detours caused by obstacles or change of intent. Our novel formulation advances Lagrangian Mechanics. 

Related to ours is prior work in cognitive science \cite{Baker2006} aimed at inferring human goal destinations based on inverting a probabilistic generative model of goal-dependent plans from an incomplete sequence of human behavior. Also, similar to our MCMC sampling, the Wang-Landau Monte Carlo (WLMC) sampling is used in \cite{Kwon2013} for people tracking in order to handle abrupt motions.

\textbf{Prediction and early decision}. 
There is growing interest in action prediction \cite{Kong2014, Lan2014, Huang2014}, and early recognition of a single human activity \cite{Ryoo2011} or a single structured event \cite{Pei2011, Hoai2012}.  These approaches are not aimed at predicting human trajectories, and are not suitable for our domain in which multiple activities may happen simultaneously. Also, some of them make the assumption that human activities are structured \cite{Pei2011, Hoai2012} which is relatively rare in our surveillance videos of public spaces where people mostly just walk or remain still. Another difference is that we distinguish activities by human intent, rather than their semantic meaning. Some early recognition approached do predict human trajectories \cite{Kim2010}, but use a deterministic vector field of people's movements, whereas our ``dark energy'' fields are stochastic. In\cite{Koppula2013}, an anticipatory temporal conditional random field (ATCRF) is used for predicting human activities based on object affordances. These activities are, however, defined at the human-body scale, and thus the approach cannot be easily applied to our wide-scene views. A linear dynamic system of \cite{Zhou2011, Zhou2012a} models smooth trajectories of pedestrians in crowded scenes, and thus cannot handle sudden turns and detours caused by obstacles, as required in our setting. In graphics, relatively simplistic models of agents are used to simulate people's trajectories in a virtual crowd \cite{Lee2007, Lerner2007, Pellegrini2012}, but cannot be easily extended to our surveillance domain. Unlike the above related work, we do not exploit appearance-based object detectors for localizing objects that can serve as possible people's destinations in the scene.

 {\bf Extensions from our preliminary work.} We extend our preliminary work \cite{ICCV13_DarkMatter} by additionally: 1) Modeling and inferring ``sequential'' human intents and ``change of intent'' along the course of people's trajectories; 2) Online prediction of human intents and trajectories; 3) Clustering functional objects; and 4) Presenting the corresponding new empirical results. Neither change of intent nor ``sequential'' intents were considered in \cite{ICCV13_DarkMatter}.
 
\subsection{Contributions}
This paper makes the following three contributions.
\begin{itemize}
\item {\em Agent-based Lagrangian Mechanics (ALM).} We leverage the Lagrangian mechanics (LM) for modeling human motion in an outdoor scene as a physical system. The LM is extended to account for human free will to choose goal destinations and change intent. 

\item {\em Force-dynamic functional map.} We present a novel approach to modeling and estimating the force-dynamic functional map of a scene in the surveillance video. 

\item {\em Human intents.} We explicitly model latent human intents, and allow a person to change intent.
\end{itemize}


\section{Agent-based Lagrangian Mechanics}
\label{sec:background}
At the scale of large scenes such as courtyard, people are considered as ``particles'' whose shapes and dimensions are neglected, and their motion dynamics modeled within the framework of Lagrangian mechanics (LM). LM studies the motion of a particle with mass, $m$, at positions ${\bf x}(t) = (x(t), y(t))$ and velocity, $ \dot{{\bf x}}(t)$, in time $t$,  in a force field $\vec{F}({\bf x}(t))$ affecting the motion of the particle. Particle motion in generalized coordinates system is determined by the Lagrangian function, $L({\bf x}, \dot{{\bf x}}, t)$, defined as the kinetic energy of the entire physical system, $\frac{1}{2}m\dot{{\bf x}}(t)^2$, minus its potential energy, $-\int_{\bf x} \vec{F}({\bf x}(t))\vec{d{\bf x}}(t)$, 
\begin{equation}
L({\bf x}, \dot{{\bf x}}, t) =  \frac{1}{2}m\dot{{\bf x}}(t)^2 + \int_{\bf x} \vec{F}({\bf x}(t))\vec{d{\bf x}}(t). 
\end{equation}
Action in such a physical system is defined as the time integral of the Lagrangian of trajectory ${\bf x}$ from $t_1$ to $t_2$: 
\begin{equation}
\text{Action}=\int_{t_1}^{t_2} L(\mathbf{x}, \dot{\mathbf{x}}, t) dt. 
\end{equation}
LM postulates that a particle's trajectory, $\Gamma(t_1, t_2)=[{\bf x}(t_1),...,{\bf x}(t_2)]$, is governed by the principle of Least Action in a generalized coordinate system: 
\begin{equation}
\Gamma(t_1, t_2) = \arg \min_{\mathbf{x}}\int_{t_1}^{t_2} L(\mathbf{x}, \dot{\mathbf{x}}, t) dt.
\end{equation}

\begin{figure*}
\centering
\includegraphics[width=0.8\linewidth]{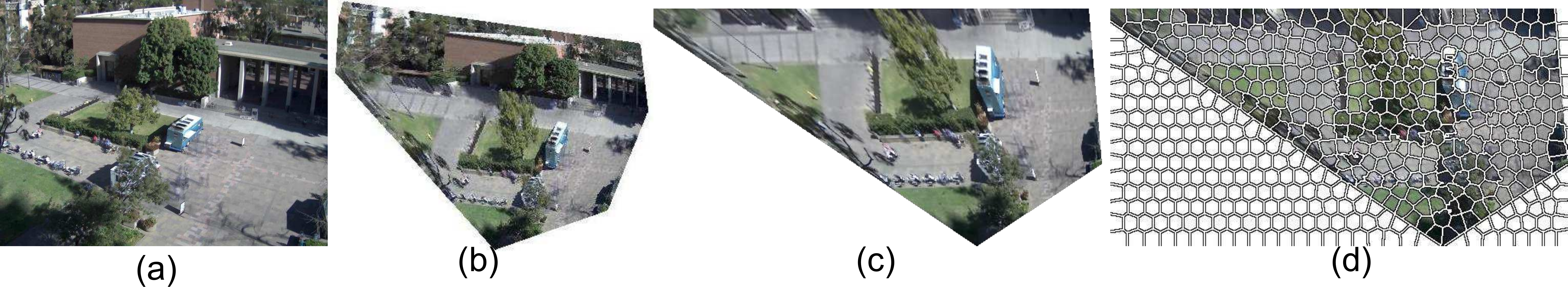}
\caption{(a) An example of a public space; (b) 3D reconstruction of the scene using the method of \cite{Zhao2011}; (c) Our estimation of the ground surface; and (d) Our inference is based on superpixels obtained using the method of \cite{Achanta2012}.}
\label{fig:segmentation}
\end{figure*}

The classical LM is not directly applicable to our domain, because it considers inanimate objects. We extend LM in two key aspects, and thus derive the Agent-based Lagrangian mechanics (ALM). In ALM, a physical system consists of a set of force sources. Our first extension enables the particles to become agents with free will to select a particular force source from the set which can drive their motion. Our second extension endows the agents with knowledge about the layout map of the physical system. Consequently, by the principle of Least Action, they can globally optimize  their shortest paths toward the selected force source, subject to the known layout of obstacles. These two extensions can be formalized as follows.

Let $i$th agent choose $j$th source from the set of sources. Then, $i$'s action, i.e., trajectory is
\begin{equation}
\begin{array}{l}
\Gamma_{ij}( t_1, t_2) \\
{=} \displaystyle \arg \min_{\mathbf{x}} \int_{t_1}^{t_2} \Big[ \frac{1}{2}m\dot{\mathbf{x}}(t)^2 {+}\int_\mathbf{x} \vec{F}_{ij}(\mathbf{x}(t))\vec{d\mathbf{x}}(t)\Big]dt,\\
\quad  \text{s.t.}\quad  \mathbf{x}(t_1) = \mathbf{x}_i, \; \mathbf{x}(t_2) = \mathbf{x}_j.
\end{array}
\label{trajectory_least_action1}
\end{equation}
For solving the difficult optimization problem of (\ref{trajectory_least_action1}) we resort to certain approximations, as explained below.
 
In our domain of public spaces, the agents cannot increase their speed without limit. Hence, every agent's speed is upper bounded by some maximum speed. Also, it seems reasonable to expect that accelerations or decelerations of people along their trajectories in a public space span negligibly  short time intervals. Consequently, the first term in (\ref{trajectory_least_action1}) is assumed to depend on a constant velocity of the agent, and thus does not affect estimation of $\Gamma_{ij}( t_1, t_2)$. 


For simplicity, we allow the agent to make only discrete displacements over a lattice of scene locations $\Lambda$ (e.g., representing centers of superpixels occupied by the ground surface in the scene), i.e., $\vec{d\mathbf{x}}(t) = \vec{\Delta\mathbf{x}}$. Also, we expect that the agent is reasonable and always moves along the direction of $\vec{F}_{ij}(\mathbf{x})$ at every location.

From (\ref{trajectory_least_action1}) and above considerations, we derive:
\begin{equation}
\Gamma_{ij}(t_1, t_2) =\arg\min_{\Gamma\subset\Lambda} \displaystyle \sum_{{\bf x}\in \Gamma} |\vec{F}_{ij}({\bf x})\cdot \vec{\Delta{\bf x}}|\\
\label{Gamma_estimation}
\end{equation}
such that $\mathbf{x}(t_1) = \mathbf{x}_i$ and $\mathbf{x}(t_2) = \mathbf{x}_j$.

A globally optimal solution of (\ref{Gamma_estimation}) can be found with the Dijkstra algorithm. Note that the end location of the predicted $\Gamma_{ij}(t_1, t_2)$ corresponds to the location of source $j$. It follows that estimating human trajectories can readily be used for estimating the functional map of the scene. To address uncertainty, this estimation is formulated within the Bayesian framework, as explained next.
 

\section{Problem Formulation}
\label{sec:formulation}
This section defines observable and latent variables of our framework in a ``bottom-up'' way. These definitions will be used in Section \ref{sec:model} for specifying the joint probability distribution of our model.

The video shows agents, $A=\{a_i:i=1,...,M\}$, and sources of ``dark energy'', $S=\{{\bf s}_j:j=1,...,N\}$, occupying locations on the ground represented by a 2D lattice, $\Lambda=\{{\bf x}=(x,y):x,y\in \mathbb{Z}_+\}$. The locations ${\bf x}\in\Lambda$ may be walkable or non-walkable, as indicated by a constraint map, $C=\{c({\bf x}):\forall {\bf x}\in \Lambda,~c({\bf x})\in\{-1,1\}\}$, where $c({\bf x})=-1$, if ${\bf x}$ is non-walkable, and $c({\bf x})=1$, otherwise. Locations in scene where agents may occur form the set  $\Lambda_1=\{{\bf x}:  {\bf x}\in \Lambda, ~c({\bf x}) {=} 1\}$. Below, we define the prior probabilities and likelihoods of these variables that are suitable for our setting. 


\subsection{Constraint map}
{\bf Smoothness prior distribution.} The prior $P(C)$ enforces spatial smoothness using the Ising random field: 
\begin{equation}
P(C) {\propto}\textstyle\exp\left[\beta \displaystyle \sum_{{\bf x}\in \Lambda, {\bf x}' \in \partial{\bf x}\cap \Lambda} c({\bf x})c({\bf x}')\right], ~\beta>0.
\label{eqn:prior_C}
\end{equation}

{\bf Video appearance features.} We are also interested in modeling walkable surfaces in our surveillance videos in terms of appearance features. The likelihood of these appearance features is defined as 
\begin{equation}
P(I|C) = \prod_{{\bf x}\in\Lambda} P(\phi({\bf x})|c({\bf x}){=}1),
\label{eqn:appearance}
\end{equation}
where $\phi({\bf x})$ is a feature descriptor vector consisting of: i) the RGB color at the scene location ${\bf x}$, and ii) the binary indicator if ${\bf x}$ belongs to the ground surface of the 3D reconstructed scene. $P(\phi({\bf x})|c({\bf x}){=}1)$ is specified as a 2-component Gaussian mixture model. Note that $P(\phi({\bf x})|c({\bf x}){=}1)$ is directly estimated on our given (single) video with latent $c({\bf x})$, not using training data.

\subsection{Dark Matter}
The latent functional object, ${\bf s}_j \in S$, are characterized by
\begin{equation}
S=\{{\bf s}_j=(\bm{\mu}_j, \Sigma_j):j=1,...,N\}, 
\end{equation}
where $\bm{\mu}_j \in \Lambda$ is the location of ${\bf s}_j$, and $\Sigma_j$ is a $2\times2$ spatial covariance matrix of ${\bf s}_j$'s force field. The distribution of $S$ is conditioned on $C$, where the total number $N=|S|$ and occurrences of the sources are modeled with the Poisson and Bernoulli pdf's:
\begin{equation}
P(S|C) {\propto}\frac{\eta^N}{N!}e^{-\eta} \prod_{j=1}^{N}\rho^{\frac{c(\bm{\mu}_j)+1}{2}}(1-\rho)^{\frac{1-c(\bm{\mu}_j)}{2}}
\label{eqn:prior_S}
\end{equation}
where parameters $\eta>0$, $\rho\in(0,1)$, and $c(\bm{\mu}_j)\in\{-1,1\}$\\

\subsection{Human Intents}
People's intents are specified by the set of agent-goal relationships $R=\{r_{ij}\}$. Specifically, when agent $a_i$ wants to pursue ${\bf s}_j$, we specify their relationship as $r_{ij}= 1$; otherwise $r_{ij} = 0$. As explained in the following subsection, note that $a_i$ may want to pursue more than one source from $S$ in a sequential manner, during the lifetime of the trajectory. Consequently, we may have $\sum_j r_{ij}\ge 1$.

The distribution of $R$ is conditioned on $S$, and modeled using the multinomial distribution with parameters $\bm{\theta}=[\theta_1,...,\theta_j,...,\theta_{N}]$, 
\begin{equation}
P(R|S)=\textstyle\prod_{j=1}^{N}\theta_j^{b_j},
\label{eqn:prior_R}
\end{equation}
where each $\theta_j$ is viewed as a prior of selecting ${\bf s}_j\in S$, and ${\bf s}_j$ is chosen $b_j$ times  to serve as a goal destination,  $b_j= \sum_{i=1}^M \mathbbm{1}(r_{ij} = 1)$, $j=1,...,N$, where $\mathbbm{1}(\cdot)$ denotes the binary indicator function. 

\subsection{Three Types of Intent Behavior}\label{sec:intent_behavior}
In this paper, we consider three types of intent behavior of an agent -- namely, ``single'', ``sequential'' and ``change of intent''. The intent behavior of all agents is represented by a set of latent variables, $Z=\{z_i\}$. We make the simplifying assumption that $R$ and $Z$ are independent, and that individual intent behavior $z_i$ is independent from other agents' behaviors:
\begin{equation}
P(Z) = \prod_{i=1}^M P(z_i),
\label{behavior}
\end{equation}
where $P(z_i)$ is specified below.

\subsubsection{Single Intent}
An agent $a_i$ is assigned  $z_i=\text{``single''}$ when its intent is to achieve exactly one goal, and remain at the reached destination indefinitely. E.g., a ``single intent'' agent may intend to exit the field of view and never return to the scene. The probability that an agent has ``single intent'' is 
\begin{equation}
P(z_i=\text{``single''})\propto1-\kappa,\quad \kappa\in[0,1].
\label{eq:simple}
\end{equation}

Note that if $z_i=\text{``single''}$ we have $r_{ij}=1$ only for a single source, and  is piecewise constant in $t$, but may vary over time.

\subsubsection{Sequential Intents}
An agent $a_i$ is assigned  $z_i=\text{``sequential''}$ when its intent is to achieve several goals along the trajectory. Each goal in the sequence serves as a milestone, where the person satisfies the corresponding need before moving to the next goal. E.g., someone may walk to a vending machine to buy a drink, and then walk toward a bench to get some rest. The probability that an agent has $n$ ``sequential intents'' is specified as 
\begin{equation}
P(z_i=\text{``sequential''}, n) \propto \kappa^{(n-1)}(1-\kappa),\quad \kappa\in[0,1].
\label{eq:complex}
\end{equation}
In our videos, we have $2 \leq n \leq 3$.

\subsubsection{Change of Intent}

An agent may also give up on the initial goal before reaching it, and switch to another goal. This defines the third type of intent behavior, called ``change of intent''. 




In our surveillance videos, we observe that ``change of intent'' happens relatively seldom, on average once per person. Therefore, we make the assumption that an agent can change the intent only once, and that moment may happen at any time between the start and end of the trajectory with probability $\gamma\in[0,1]$. Also, we specify that the new goal can be selected from the remaining $N-1$ possible destinations in the scene with a uniform prior distribution. Hence, the probability of ``change of intent'' is defined as
\begin{equation}
P(z_i=\text{``change ''})
= \frac{\gamma}{N-1}.
\label{eq:change}
\end{equation}

\subsection{Forces}\label{sec:forces}
Functional objects in the scene exert either repulsive or attractive forces on the agents.  

{\bf Repulsion Forces.}  Every non-walkable location ${\bf x}'\in\Lambda{\setminus}\Lambda_1$ generates a repulsion force at agent's location ${\bf x}$, $\vec{F}^{-}_{{\bf x}'}({\bf x})$. The magnitude $|\vec{F}^{-}_{{\bf x}'}({\bf x})|$ is defined as a Gaussian function of the quadratic $({\bf x}-{\bf x}')^2$, with covariance $\Sigma = \sigma_r^2 \mathcal{I}$, where $\mathcal{I}$ is the identity matrix, and $\sigma_r^2 = 10^{-2}$ is empirically found as best. Thus, the magnitude $|\vec{F}^{-}_{{\bf x}'}({\bf x})|$ is large in the vicinity of non-walkable location ${\bf x}'$, but quickly falls to zero for locations farther away from ${\bf x}'$. This models our observation that a person may take a path that is very close to non-walkable areas, i.e., the repulsion force has a short-range effect on human trajectories. The sum of all repulsion forces arising from non-walkable areas in the scene gives the joint repulsion, $\vec{F}^{-}({\bf x})=\sum_{{\bf x}'\in\Lambda\setminus\Lambda_1}\vec{F}^{-}_{{\bf x}'}({\bf x})$.


{\bf Attraction Forces.} Each source ${\bf s}_j\in S$ is capable of generating an attraction force, $\vec{F}^{+}_j({\bf x})$, if selected as a goal destination by an agent at location ${\bf x}$. The magnitude $|\vec{F}_{j}^+({\bf x})|$ is specified as a Gaussian function of the quadratic $({\bf x}-{\bf x}_j)^2$, with 
covariance $\Sigma = \sigma_a^2 \mathcal{I}$ taken to be the same for all sources ${\bf s}_j\in S$, and $\sigma_a^2 = 10^{4}$ is empirically found as best. This models our observation that people tend to first approach near-by functional objects, because, in part, reaching them requires less effort than approaching farther destinations in the scene. The attraction force is similar to the gravity force in physics whose magnitude becomes smaller as the distance increases.

When $a_i\in A$ selects ${\bf s}_j\in S$, $a_i$ is affected by the cumulative force, $\vec{F}_{ij}({\bf x})$, estimated as:
\begin{equation}
\vec{F}_{ij}({\bf x})= \vec{F}^{+}_j({\bf x}) + \vec{F}^{-}({\bf x}).
\label{force} 
\end{equation}
Note that an instantiation of latent variables $C,S,R$ uniquely defines the force field $\vec{F}_{ij}({\bf x})$ in (\ref{force}). 

\begin{figure}
\centering
\includegraphics[width=0.8\columnwidth]{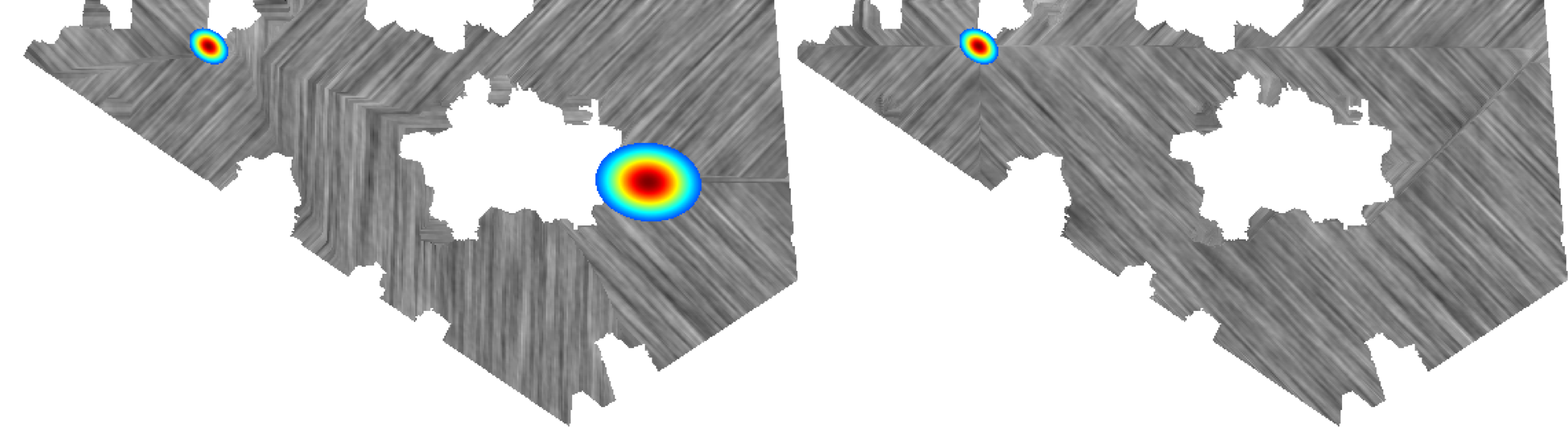}\hfill%
\caption{Visualizations of the force field for the scene from Fig.~\ref{fig:example_scene}. ({\em left}) In LM,  particles are driven by a sum of all forces; the figure shows the resulting fields generated by only two sources. ({\em right}) In ALM, each agent selects a single force $\vec{F}_{j}({\bf x})$ to drive its motion; the figure shows that forces at all locations in the scene point toward the top left of the scene where the source is located. The white regions represent our estimates of obstacles. Repulsion forces are short ranged, with magnitudes too small to show here.}
\label{fig:difference}
\end{figure}

From above, we can more formally specify the difference between LM and our ALM. In LM, an agent would be affected by a sum of forces of all sources in $S$, $\sum_{j} \vec{F}^{+}_j({\bf x})$, along with the joint repulsion force. In contrast, in ALM, an agent is affected by the force of a single selected source, $\vec{F}_{j}({\bf x})$, along with the joint repulsion force. The difference between between LM and our ALM is illustrated in Fig.~\ref{fig:difference}. 

\subsection{Trajectories}
Recall that Sec.~\ref{sec:background} formalizes that an optimal trajectory $\Gamma_{ij}(t_1,t_2)=[{\bf x}(t_1)={\bf x}_{i},\dots,{\bf x}(t_2)={\bf x}_{j}]$ of $a_i$ at location ${\bf x}_{i}$ moving toward ${\bf s}_j$ at location ${\bf x}_{j}$ minimizes the energy $\sum_{{\bf x}\in \Gamma_{ij}} |\vec{F}_{ij}({\bf x})\cdot\vec{\Delta{\bf x}}|$. Simplifying notation, we extend this formulation to account the agent's intent behavior as 
\begin{equation}
\Gamma_{i} = \sum_j \Gamma_{ij} = \arg\min_{\Gamma\subset\Lambda} \sum_j  \sum_{{\bf x}\in \Gamma} |\vec{F}_{ij}({\bf x})\cdot \vec{\Delta{\bf x}}|
\label{multiple_goals_G}
\end{equation}
where the summation over $j$ uses: (i) only one source for ``single'' intent (i.e., $\Gamma_{i}= \Gamma_{ij}$ when ${r_{ij}=1}$), (ii) two sources for ``change of intent'', and (iii) maximally $n$ sources for ``sequential" intent. Also the minimization in (\ref{multiple_goals_G}) is constrained such that the trajectory must {\em sequentially} pass through locations ${\bf x}_j$ of all sources ${\bf s}_j$ pursued for ``sequential'' intents.

To address the uncertainty about the layout of obstacles, agent's true goal destination(s), and agent's intent behavior, we specify the likelihood of $\Gamma_{i}$ in terms of the energy that  $a_i$ must spend moving along the trajectory as
\begin{equation}
\setlength{\arraycolsep}{2pt}
\begin{array}{l}
P(\Gamma_{i}|C,S, R,z_i) \\
{\propto} \exp\left[-\lambda  \sum_j\sum_{{\bf x}\in \Gamma_{ij}} |\vec{F}_{ij}({\bf x})\cdot \vec{\Delta{\bf x}}|\right],
 \end{array}
\label{likelihood_G}
\end{equation}
where $\lambda >0$, and $R$ specifies the source(s) ${\bf s}_j$ that $a_i$ is (sequentially) attracted to. The likelihood in (\ref{likelihood_G}) models that when $a_i$ is far away from ${\bf s}_j$, the total energy needed to cover that trajectory is bound to be large, and consequently uncertainty about $a_i$'s trajectory is large. Conversely, as $a_i$ gets closer to ${\bf s}_j$, uncertainty about the trajectory reduces. Note that applying the principle of Least Action to (\ref{likelihood_G}), similar to (\ref{multiple_goals_G}), gives the highest likelihood of $\Gamma_{i}$. 

\section{The Probabilistic Model}
\label{sec:model}
This section defines the joint posterior distribution of latent random variables  $W=\{C, S, R, Z, \Gamma\}$ given the initially observed human trajectories $\Gamma'=\{\Gamma'_{i}\}$ and appearance features $I$ in the video as%
\begin{equation}
P(W|\Gamma', I)
\propto P(C, S, R, Z) P( \Gamma,\Gamma', I | C, S, R, Z),
 \label{eqn:joint_posterior}
\end{equation}
where
\begin{equation}
P(C, S, R, Z) = P(C)P(S|C)P(R|S)P(Z), 
\label{eqn:joint_posterior_expansion_1}
\end{equation}
with distributions $P(C)$, $P(S|C)$, $P(R|S)$ and $P(Z)$ defined in (\ref{eqn:prior_C}), (\ref{eqn:prior_S}), (\ref{eqn:prior_R}) and (\ref{behavior}), respectively,
\begin{equation}
P( \Gamma,\Gamma', I | C, S, R, Z) = P(\Gamma,\Gamma' |C, S, R, Z)P(I|C), 
\label{eqn:joint_posterior_expansion_2}
\end{equation}
with distribution $P(I|C)$  defined in (\ref{eqn:appearance}),
and
\begin{equation}
P(  \Gamma,\Gamma'| C, S, R, Z) = \displaystyle
\prod_{i=1}^M P(\Gamma_{i}|C,S,R, Z),
\label{eqn:joint_posterior_expansion_3}
\end{equation}
where $P(\Gamma_{i}|C,S,R, Z)$ is given by (\ref{likelihood_G}). Note that in (\ref{eqn:joint_posterior_expansion_3})  $\Gamma_{i}(0,t_0)=\Gamma'_{i}(0,t_0)$ over the initial time interval $(0,t_0)$ when we are allowed to observe the video, and $\Gamma_{i}(t_0,T)$ represents a trajectory of $a_i$ in the future unobserved time interval $(t_0,T)$.

\section{Offline and Online Inference}
\label{sec:inference}

Given observations $\{I, \Gamma'\}$ in the time interval $(0,t_0)$, we infer the latent variables $W$ in the future time interval $(t_0,T)$ by maximizing the joint posterior defined in (\ref{eqn:joint_posterior}). We consider offline and online inference. 

Offline inference first estimates $C$, $S$, and $R$ over the initial observable time interval $(0,t_0)$. These estimates are then used to compute forces $\{\vec{F}_{ij}({\bf x})\}$ for all agents and their respective goal destinations, as specified in Sec.~\ref{sec:forces}. Finally, the computed forces are used to predict the entire trajectories of agents over $(t_0,T)$ using the Dijkstra algorithm, and estimate the agents' intent behaviors. 

In online inference, the agent-source relationships $R^{(t)}$ and trajectories $\Gamma^{(t)}=\Gamma(0,t)$ are sequentially predicted frame by frame. Thus, new evidence about the agent-source relationships, provided by previous trajectory predictions up to frame $t$, is used to re-estimate $R^{(t+1)}$. This re-estimation, in turn, is used to predict $\Gamma^{(t+1)}$ in the next frame $(t+1)$. 

Note that in offline inference we seek to infer all three types of intent behavior for each agent. In online inference, however, we do not consider ``change of intent'', because this would require an explicit modeling of the statistical dependence between $R$ and $Z$, and transition probabilities between $R^{(t)}$ and $R^{(t+1)}$ which is beyond our scope.

In the following, we first describe the data-driven MCMC process \cite{Kwon2013, Tu2002} used for estimating $C$, $S$, and $R$. Then, we present our sequential estimation of the agents' trajectories for online inference.

\subsection{Scene Interpretation}\label{sec:Scene_Interpretation}

\begin{figure}[t]
\centering
\includegraphics[width=0.8\columnwidth]{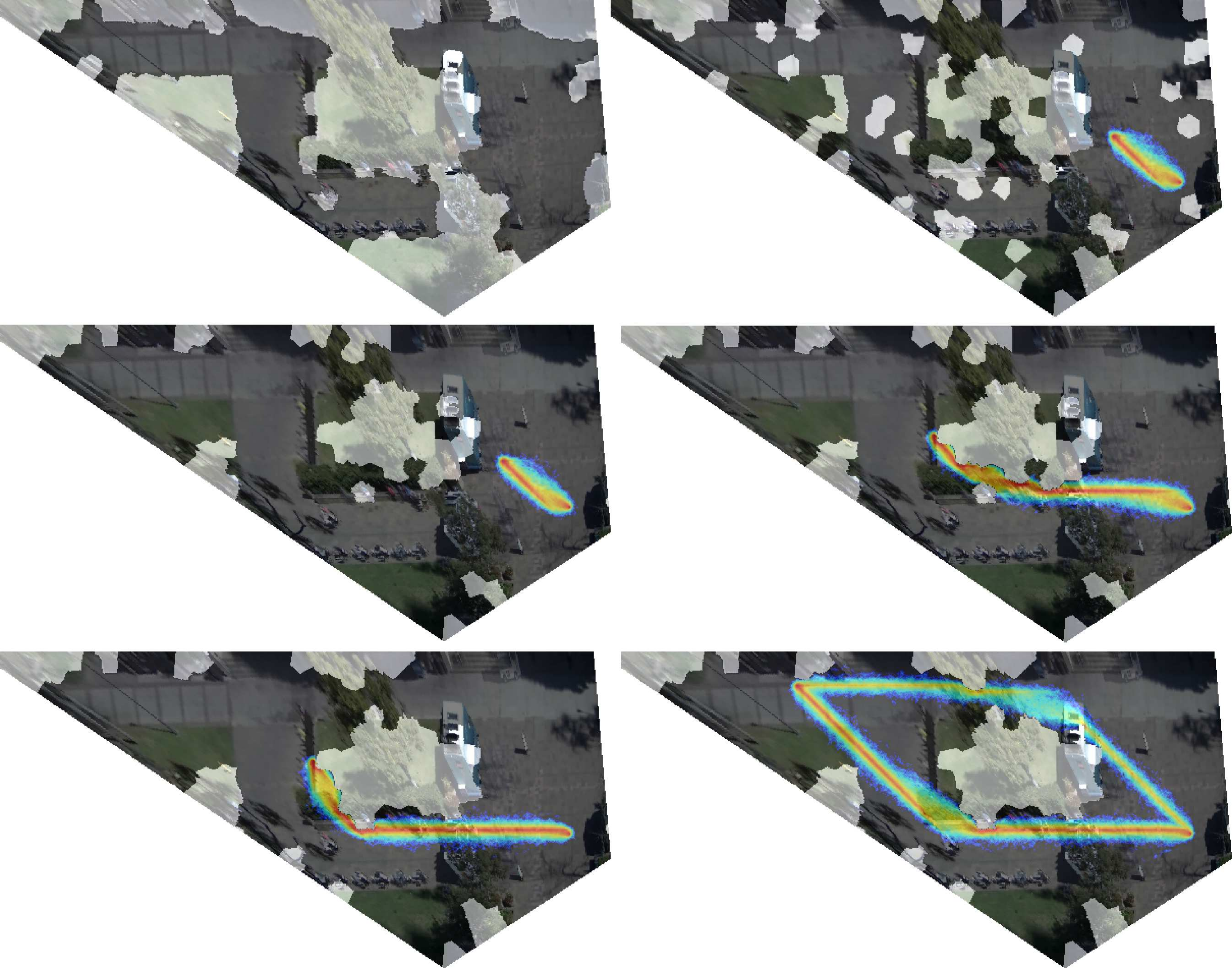}
\caption{Top view of the scene from Fig.~\ref{fig:example_scene} with the overlaid illustration of the MCMC inference. The rows show the progression of proposals of the constraint map $C$ in raster scan  (the white regions indicate obstacles), and trajectory estimates of agent $a_i$ with goal to reach ${\bf s}_j$ (warmer colors represent higher likelihood $P(\Gamma_{ij}| C, S, r_{ij}=1, z_i=\text{``single''})$). In the last iteration (bottom right), MCMC estimates that the agent's goal is to approach ${\bf s}_j$ at the top-left of the scene, and finds two equally likely trajectories for this goal.} 
\label{fig:MCMC}
\end{figure}

\begin{figure}[t]
\centering
\includegraphics[width=0.8\linewidth]{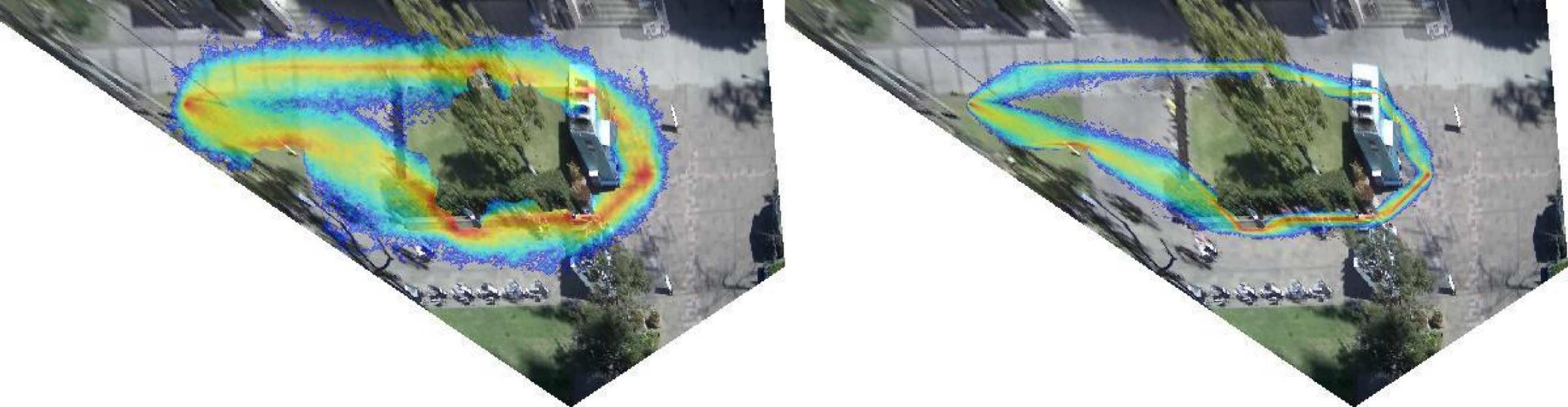}
\caption{Top view of the scene from Fig.~\ref{fig:example_scene} with the overlaid trajectory predictions of a person who starts at the top-left of the scene, and wants to reach the dark matter in the middle-right of the scene (the food truck).  A magnitude of difference in parameters $\lambda=0.2$ ({\em on the left})  and  $\lambda=1$ ({\em on the right}) used to compute likelihood $P( \Gamma_{ij}| C, S, R, Z)$ gives similar trajectory predictions. The predictions are getting more certain as the person comes closer to the goal. Warmer colors represent higher likelihood.}
\label{fig:TwoTrajectories}
\end{figure}

To estimate $C$, $S$, and $R$ over interval $(0,t_0)$ when the scene in the video can be observed, we use a data-driven MCMC \cite{Kwon2013, Tu2002}, as illustrated in Figures~\ref{fig:MCMC}~and~\ref{fig:TwoTrajectories}. MCMC provides theoretical guarantees of convergence to the optimal solution. 

Each step of our MCMC proposes a new solution $Y_\text{new}{=}\{C_\text{new},S_\text{new},R_\text{new}\}$. The decision to discard the current solution, $Y{=}\{C,S,R\}$, and accept $Y_\text{new}$ is made based on the acceptance rate, 
\begin{equation}
\alpha=\min(1,\frac{Q(Y{\to} Y_\text{new})}{Q(Y_\text{new}\to Y)}\frac{P(Y_\text{new}| \Gamma', I)}{P(Y| \Gamma', I)}).
\label{eq:alpha}
\end{equation}
If $\alpha$ is larger than a threshold uniformly sampled from $[0,1]$, the jump to $Y_\text{new}$ is accepted. In (\ref{eq:alpha}),
the proposal distribution is defined as 
\begin{equation}
Q(Y{\to} Y_\text{new}) = Q(C{\to} C_\text{new})Q(S{\to} S_\text{new})Q(R{\to} R_\text{new}),
\end{equation}
and the posterior distribution 
\begin{equation}
P(Y| \Gamma', I)\propto P(C, S, R) P( \Gamma', I | C, S, R, Z').
\label{eq:posteriorY}
\end{equation}
Each term in (\ref{eq:posteriorY}) is already specified in (\ref{likelihood_G}), (\ref{eqn:joint_posterior_expansion_1}), (\ref{eqn:joint_posterior_expansion_2}) and (\ref{eqn:joint_posterior_expansion_3}). Note that in (\ref{eq:posteriorY}) we make the assumption that the observation interval $(0,t_0)$ is too short for agents to exhibit more complex intent behaviors beyond ``single intent''. Therefore, for all agents we set  that their initial $Z'=\{z'_i=\text{``single''}\}$ in $(0,t_0)$. Also, note that the prior $P(Z')$ gets canceled out in the ratio in (\ref{eq:alpha}), and hence $P(Z')$ is omitted in (\ref{eq:posteriorY}).

The initial $C$  is proposed by setting $c({\bf x}) = 1$ at all locations covered by observed trajectories $\Gamma'$, and randomly setting $c({\bf x}) = -1$ or $c({\bf x}) = 1$ for other locations. The initial number $N$ of sources in $S$ is probabilistically sampled from the Poisson distribution of (\ref{eqn:prior_S}), while their layout is estimated as $N$ most frequent stopping locations in $\Gamma'$. Given $\Gamma'$ and $S$, we probabilistically sample the initial $R$ using the multinomial distribution in (\ref{eqn:prior_R}). In the subsequent MCMC iterations, new solutions $C_\text{new}$, $S_\text{new}$, and $R_\text{new}$ are sequentially proposed and accepted as current solutions based on the acceptance rate $\alpha$.

{\bf The Proposal of $C_\text{new}$} randomly chooses ${\bf x}\in \Lambda$, and reverses its polarity, $c_\text{new}({\bf x})=-c({\bf x})$. The proposal distribution $Q(C {\rightarrow} C_\text{new})=Q(c_\text{new}({\bf x}))$ is data-driven. $Q(c_\text{new}({\bf x})=1)$ is defined as the normalized average speed of people observed at $\bf x$, and $Q(c_\text{new}({\bf x})=-1) = 1- Q(c_\text{new}({\bf x})=1)$.
 
{\bf The Proposal of $S_\text{new}$} includes the ``death'' and ``birth'' jumps.  The birth jump randomly chooses ${\bf x}\in \Lambda_1$, and adds a new source ${\bf s}_{N+1} = (\bm{\mu}_{N+1},\Sigma_{N+1})$ to $S$, resulting in $S_\text{new}=S\cup\{{\bf s}_{N+1}\}$, where $\bm{\mu}_{N+1}={\bf x}$, and $\Sigma_{N+1} = \text{size}^2\mathcal{I}$, where $\text{size}$ is the scene size (in pixels). The death jump randomly chooses an existing source ${\bf s}_j\in S$, and removes it from $S$,  resulting in $S_\text{new}=S\setminus\{{\bf s}_{j}\}$. The ratio of the proposal distributions is specified as $\frac{Q(S {\rightarrow} S_\text{new})}{Q(S_\text{new} {\rightarrow} S)}=1$, 
indicating no preference to either `death'' or ``birth'' jumps. That is, the proposal of $S_\text{new}$ is governed by the Poisson prior of (\ref{eqn:prior_S}), and trajectory likelihoods $P( \Gamma'| C, S, R, Z')$, given by (\ref{likelihood_G}), when computing the acceptance rate $\alpha$.

{\bf The Proposal of $R_\text{new}$} randomly chooses one person $a_i\in A$ with goal ${\bf s}_j$, and performs one of the three possible actions: (i) randomly changes  $a_i$'s goal to ${\bf s}_k\in S$, (ii) randomly adds a new goal ${\bf s}_k\in S$ to $a_i$ if $\sum_j r_{ij}<n$ where $n$ is the maximum number of goals (in our domain $n=3$), and (iii) randomly removes one of current goals of $a_i$ if $\sum_j r_{ij}>1$. The changes in the corresponding relationships $r_{ij}\in R$ result in $R_\text{new}$. The ratio of the proposal distributions is $\frac{Q(R {\rightarrow} R_\text{new})}{Q(R_\text{new} {\rightarrow} R)}=1$. This means that the proposal of $R_\text{new}$ is governed by the multinomial prior $P(R|S)$ and likelihoods $P( \Gamma' | C, S, R, Z')$, given by (\ref{eqn:prior_R}) and(\ref{likelihood_G}), when computing  $\alpha$  in (\ref{eq:alpha}). 

Importantly, the random proposals of $R_\text{new}$ ensure that for every agent $a_i$ we have $\sum_j r_{ij,\text{new}}\ge1$. Since our assumption is that in $(0,t_0)$ the agents may have only a single intent, we consider that agent $a_i$ first wants to reach the closest source ${\bf s}_j$ for which $r_{ij,\text{new}}=1$, $\Gamma_i'=\Gamma'_{ij}$, out of all other sources ${\bf s}_k$ that also have $r_{ik,\text{new}}=1$ and which the agent can visit later after time $t_0$. This closest source ${\bf s}_j$  is then used to compute the likelihood $P( \Gamma'_{ij} | C, S, r_{ij}=1, z_i'=\text{``single''})$, as required in (\ref{eq:alpha}), and thus conduct the MCMC jumps.

\subsection{Offline Inference of $\Gamma$ and $Z$} \label{sec:offline_inference}

From the MCMC estimates of $C$, $S$, $R$, we readily estimate forces $\{\vec{F}_{ij}\}$, given by (\ref{force}). Then we proceed to predicting trajectories $\{\Gamma_{i}\}$ and intent behavior $\{z_i\}$ in the future interval $(t_0,T)$.

{\bf The single-intent case}. When the MCMC estimates that $a_i$ has only a single goal destination, $\sum_j r_{ij}=1$, we readily predict $z_i=\text{``single''}$, and estimate a globally optimal trajectory $\Gamma_i=\Gamma_{ij}$ using the Dijkstra algorithm. The Dijkstra path ends at location ${\bf x}_j$ of source ${\bf s}_j$ for which $r_{ij}=1$, where $a_i$ is taken to remain still until the end of the time horizon $T$.

{\bf The sequential and change of intent cases}. When the MCMC estimates that $\sum_j r_{ij}>1$, we hypothesize that the agent could have either the ``sequential'' or ``change'' of intent behavior. In this case, we jointly estimate the optimal $(\Gamma_i, z_i)^*$ pair by maximizing their joint  likelihood:
\begin{equation}
(\Gamma_i, z_i)^* =  \arg\max_{\Gamma_i,z_i} ~P( \Gamma_i| C, S, R, z_i)P(z_i),
\label{eq:joint_likelihood}
\end{equation}
for all possible configurations of $(\Gamma_i, z_i)$, where  $z_i\in\{``\text{sequential}'', ``\text{change}''\}$ and $\Gamma_i$ is the Dijkstra path spanning a particular sequence of selected sources from $S$, as explained in more detail below.

Let $J=\{j: r_{ij} = 1, j=1,\dots,N\}$ denote indices of the sources from $S$ that MCMC selected as goal destinations for $a_i$. Also, recall that for our videos of public spaces, it is reasonable to expect that people may have a maximum of $|J|\le n=3$  intents in interval $(0,T)$. The relatively small size of $J$ allows us to exhaustively consider all permutations of $J$, where each permutation uniquely identifies the Dijkstra path $\Gamma_i$ between $a_i$'s location at time $t_0$, ${\bf x}(t_0)$ and the sequence of locations ${\bf x}_j$, $j\in J$, that $a_i$ visits along the trajectory. For each permutation of $J$, and each intent behavior $z_i\in\{``\text{sequential}'', ``\text{change}''\}$, we compute the joint likelihood given by (\ref{eq:joint_likelihood}), and infer the maximum likelihood pair $(\Gamma_i, z_i)^*$.

\subsection{Online Inference of $\Gamma^{(t)}$ and $R^{(t)}$}
In online inference, we first estimate $C$, $S$, and $R^{(t_0)}$ over the time interval $(0,t_0)$ using the data-driven MCMC as described in Sec.~\ref{sec:Scene_Interpretation}. Then, we compute forces $\{\vec{F}_{ij}^{(t_0)}({\bf x})\}$ for all agents and their respective goal destinations identified in $R^{(t_0)}$. Also, we take the observed trajectories $\Gamma'$ as initial trajectory estimates, $\Gamma_{i}^{(t_0)}=\Gamma_{i}(0,t_0)=\Gamma'_i$, $i=1,\dots,M$. This initializes our online inference of $\Gamma^{(t)}$ and $R^{(t)}$ at times $t\in(t_0,T)$. Then, $\Gamma^{(t)}$ are taken to provide new cues for predicting $R^{(t+1)}$ using the MCMC described in Sec.~\ref{sec:Scene_Interpretation}. Any updates in $R^{(t+1)}$ are used to compute $\{\vec{F}_{ij}^{(t+1)}({\bf x})\}$, and subsequently update $\Gamma^{(t+1)}$ as follows. 

Recall that in online inference we do not consider ``change of intent'' behavior. In case $R^{(t)}$ estimates that agent $a_i$ has more than one goal, $\sum_j r_{ij}^{(t)}\ge1$, our online predictions of $\Gamma_i^{(t)}$ use the heuristic that $a_i$ visits the goal destinations identified in $R^{(t)}$ in the order of how far they are from the current location of $a_i$. Consequently, predictions of $\Gamma_i^{(t)}$ become equivalent for both ``single'' and ``sequential'' intent behavior, since $a_i$ always wants to reach the closest goal destination first, i.e., $\Gamma_i^{(t)}=\Gamma_{ij}^{(t)}$ for $r_{ij}^{(t)}=1$ and ${\bf x}_j$ is the closest location to $a_i$ at time $t$. From (\ref{likelihood_G}), it is straightforward to derive the conditional likelihood of $\Gamma_{ij}^{(t+1)}$, given $\Gamma_{ij}^{(t)}$ and $R^{(t)}$ as
\begin{equation}
\setlength{\arraycolsep}{2pt}
\begin{array}{l}
 P(\Gamma_{ij}^{(t+1)}|C,S, r_{ij}^{(t)}=1, z_i=\text{``single''}, \Gamma_{ij}^{(t)}) \\
 \propto \exp\Big[-\lambda \Big(|\vec{F}^{(t)}_{ij}|\cdot|{\bf x}^{(t+1)} - {\bf x}^{(t)}|
\\\quad\quad\quad\quad\quad\quad
+\min_{{\bf x}}\sum_{{\bf x}={\bf x}^{(t+1)}}^{{\bf x}_j} |\vec{F}^{(t)}_{ij}({\bf x})\cdot \vec{\Delta{\bf x}}|\Big)\Big],
 \end{array}
\label{likelihood_G2}
\end{equation}
where the second term in (\ref{likelihood_G2}) represent the energy that $a_i$ needs to spend while walking along the Dijkstra path from ${\bf x}^{(t+1)}$ to the goal destination ${\bf x}_j$.

Our online inference is summarized below.
\begin{itemize}
\item {\bf Input:} Observed trajectories $\Gamma' = \Gamma'(0,t_0)=\Gamma^{(t_0)}$, the MCMC estimates of $C$, $S$, $R^{(t_0)}$ computed as described in Sec.~\ref{sec:Scene_Interpretation}, and time horizon $T$. 
\item \textbf{Online trajectory prediction:} For every $a_i$ identify the closest goal destination $r_{ij}^{(t)}=1$, and compute $\Gamma_{ij}^{(t+1)} = [ \Gamma_{ij}^{(t)} , {\bf x}^{(t+1)}]$, where the next location ${\bf x}^{(t+1)}$ of the trajectory is estimated as an average of probabilistic samples $\xi$  generated from the conditional likelihood of (\ref{likelihood_G2}):
\begin{equation}
\setlength{\arraycolsep}{2pt}
\begin{array}{l}
{\bf x}^{(t+1)} = \text{MEAN}(\xi),\\
 \xi \sim \exp\Big[-\lambda \Big(|\vec{F}^{(t)}_{ij}|\cdot|\xi - {\bf x}^{(t)}|
\\\quad\quad\quad\quad\quad\quad
+\min_{{\bf x}}\sum_{{\bf x}=\xi}^{{\bf x}_j} |\vec{F}^{(t)}_{ij}({\bf x})\cdot \vec{\Delta{\bf x}}|\Big)\Big],
 \end{array}
\end{equation} 
\item \textbf{Stopping criterion:} If $\Gamma_{ij}^{(t+1)}$ exists out of the scene, agent $a_i$ has visited all goal destinations identified in $R^{(t)}$, or time reaches the horizon $t = T$.
\end{itemize}

\section{Results}\label{sec:experiments}
For evaluation, we use toy examples and real outdoor scenes. We present six types of results: (a) localization of functional objects $S$, (b) estimation of human intents $R$, (c) prediction of human trajectories $\Gamma$, (d) inference of ``single'', ``sequential'', and ``change'' intent behavior, and (e) functional object clustering. These results are computed on unobserved video parts, given access to an initial part of the video footage.  We study how our performance varies as a function of the length of the observed footage. For real scenes, note that annotating ground truth of non-walkable surfaces $C$ in a scene is difficult, since human annotators provide inconsistent subjective estimates (e.g., grass lawn can be truly non-walkable in one part of the scene, but walkable in another). Therefore, we do not quantitatively evaluate our inference of $C$. 

Note that our evaluations (a)--(d) significantly extend the work of \cite{Kitani2012} which presents results on only detecting ``exits'' and ``vehicles'' as functional objects in the scene, and predicting human trajectories for ``single intent'' that are bound to end at locations of ``exits'' and ``vehicles''. A comparison of our results for (a) with existing approaches to object detection would be unfair, since we do not have access to annotated training examples of the objects as most appearance-based methods for object recognition.

{\bf Evaluation Metrics:} For evaluating our trajectory prediction, we compute a modified Hausdorff distance (MHD) between the ground-truth trajectory $\Gamma$ and predicted trajectory $\Gamma^*$  as 
\begin{equation}
\setlength{\arraycolsep}{2pt}
\begin{array}{rcl}
\text{MHD}(\Gamma,\Gamma^*) &=& \max(d(\Gamma,\Gamma^*), d(\Gamma^*,\Gamma)),\\
d(\Gamma,\Gamma^*) &=& \frac{1}{|\Gamma|} \displaystyle \sum_{{\bf x} \in \Gamma} \min_{{\bf x}^*\in \Gamma^*}|{\bf x}-{\bf x}^*|.
 \end{array}
 \label{eqn:mhd}
\end{equation}
For comparison with \cite{Kitani2012}, we also compute the negative log-likelihood (NLL), $\log P(\Gamma_{ij}|\cdot)$, of the ground-truth trajectory   $\Gamma_{ij} (t_1,t_2)= \{{\bf x}(t_1)={\bf x}_i, \cdots, {\bf x}(t_2)={\bf x}_j \}$. From (\ref{likelihood_G}), NLL can be expressed as
\begin{equation}
\text{NLL}_P(\Gamma) = \frac{\lambda}{t_2-t_1} \sum_{t=t_1}^{t_2-1} |\vec{F}_{ij}({\bf x}(t))\cdot ({\bf x}(t+1)-{\bf x}(t))|,
\label{eqn:nll}
\end{equation}
where $\vec{F}_{ij}({\bf x}(t))$ is our estimate of the force field affecting the $i$th person in the video.

For evaluating localization of functional objects, $S$, we use the standard overlap criterion, i.e., the intersection-over-union ratio $IOU$ between our detection and ground-truth bounding box around the functional object. True positive detections are estimated for $IOU \ge 0.5$.  

For evaluating human-destination relationships, $R$, we compute a normalized Hamming distance between the ground-truth $r_{ij}$ and predicted $r_{ij}^*$ binary indicators of human-destination relationships, $\frac{\sum_j  \mathbbm{1}(r_{ij} {=} r_{ij}^*)}{ \sum_j r_{ij}}$.

Finally, for evaluating intent behavior, we use the standard classification recall and precision estimated from the confusion matrix of the three intent-behavior classes.

{\bf Baselines:}  For conducting an ablation study, we evaluate the following baselines and compare them with our full approach in the offline inference setting. The first three baselines are evaluated only on trajectories where people truly have a ``single'' intent, whereas the sixth baseline is evaluated on all trajectories. (1) ``Shortest path'' (SP) estimates the trajectory as a straight line, disregarding obstacles in the scene, given the MCMC estimates of $S$ and $R$. SP does not infer latent $C$, and in this way tests the influence of estimating $C$ on our overall performance. (2) ``Random Walk'' (RW) sequentially predicts the trajectory frame by frame, given the MCMC estimates of $C$ and $S$, where every prediction randomly selects one destination $j$ from $S$ and prohibits landing on non-walkable areas. RW does not estimate $R$, and in this way tests the influence of estimating $R$ on our overall performance. (3) ``Lagrangian Physical Move'' (PM) predicts the trajectory, given estimates of $C$ and $S$, under the sum of forces from all sources, $\vec{F}_{\text{classic}}({\bf x})=\sum_{j}\vec{F}_{ij}({\bf x})+\vec{F}^-({\bf x})$, as defined in Section~\ref{sec:formulation} for the Lagrangian Mechanics.  As RW, PM does not estimate $R$. (4) ``Greedy move'' (GM) makes the assumption that every person wants to go to the initially closest functional object, and thus sequentially predicts both the trajectory $\Gamma_i^{(t)}$ and destination $r_{ij}^{(t)}$ frame by frame, given the MCMC estimates of $C$ and $S$, where the latter is estimated by maximizing the following likelihood: 
\begin{equation}
j^*{=}\arg\max_j P(r_{ij} | \Gamma_i^{(t)}) {\propto} \exp(\tau(|{\bf x}_j{-}{\bf x}^{(t)}| {-} |{\bf x}_j{-}{\bf x}_i |)). 
\end{equation}
where ${\bf x}^{(t)}$ is the last location of $\Gamma_i^{(t)}$. This baseline also tests the merit of our MCMC estimation of $R$.

{\bf Comparison with Related Approaches.} We are not aware of prior work on estimating $S$, $R$, and $Z$  in the scene without access to manually annotated training labels of objects. We compare only with the state of the art method for trajectory prediction \cite{Kitani2012}.

{\bf Input Parameters.} In our default setting, we consider the first $50\%$ of trajectories as visible, and the remainder as unbserved. We use the following model parameters: $\beta = .05$, $\lambda = 0.5$, $\rho = 0.95$. From our experiments, varying these parameters in intervals $\beta \in [.01,.1]$, $\lambda \in [0.1, 1]$, and $\rho \in [0.85,0.98]$ does not change our results, suggesting that we are relatively insensitive to the specific choices of $\beta, \lambda,\rho$ over large intervals. $\eta$ is known. $\theta$ and $\psi$ are fitted from observed data.

\subsection{Toy Dataset}

\begin{figure}
\begin{center}
\begin{tabular}{c}   
	\includegraphics[width=0.8\columnwidth]{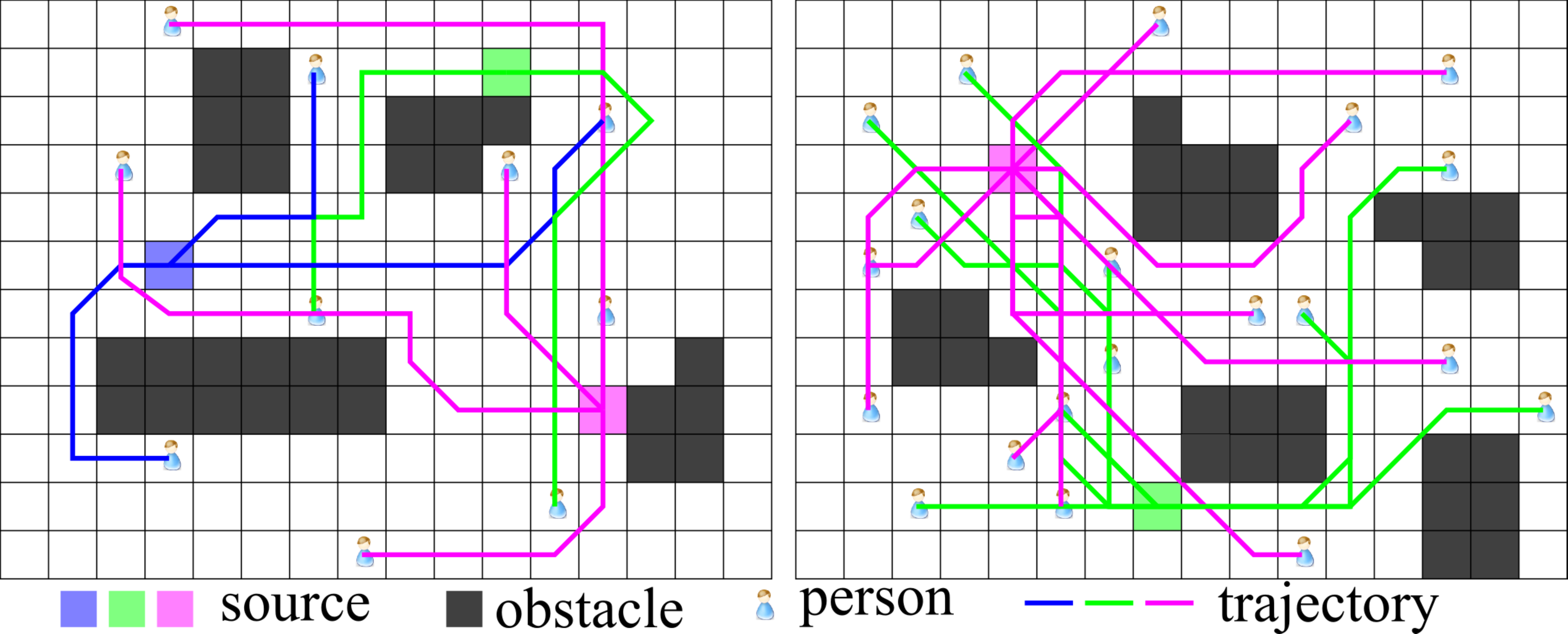}
\end{tabular}
\end{center}
\vspace{-10pt}
   \caption{Two samples from our toy dataset.}
\label{fig:toy_example}
\end{figure}

The toy dataset allows us to methodologically test our approach with respect to each dimension of the scene complexity, while fixing the other parameters. The scene complexity is defined in terms of the number of agents in the scene and the number of sources. All agents are taken to have only ``single'' intent. The scene parameters are varied to synthesize the toy artificial scenes in a rectangle random layout, where the ratio of obstacle pixels over all pixels is about $15\%$.  We vary $|S|$ and $|A|$, and we generate 3 random scene layouts for each setting of parameters $|S|$ and $|A|$. Fig.~\ref{fig:toy_example} shows two examples from our toy dataset. For inference, we take the initial $50\%$ of trajectories as observed.  Tab. \ref{tab:res_toy} shows that our approach can handle large variations in each dimension of the scene complexity.

\begin{table}\scriptsize
\begin{center}
\begin{tabular}{|c|c|c|c|c|c|c|c|c|}
\hline
\multirow{2}{*}{$|S|$} & \multicolumn{4}{|c|}{$S\&R$} & \multicolumn{4}{|c|}{NLL}  \\
\cline{2-9} 
& 10 & 20 & 50 & 100 & 10 & 20 & 50 & 100  \\ \hline
2 & 0.95 & 0.97 & 0.96 & 0.96 & 1.35 & 1.28 & 1.17 & 1.18 \\ \hline
3 & 0.87 & 0.90 & 0.94 & 0.94 & 1.51 & 1.47 & 1.35 & 1.29 \\ \hline
5 & 0.63 & 0.78 & 0.89 & 0.86 & 1.74 & 1.59 & 1.36 & 1.37 \\ \hline
8 & 0.43 & 0.55 & 0.73 & 0.76 & 1.97 & 1.92 & 1.67 & 1.54 \\ \hline
\end{tabular}
\caption{Accuracy of $S$ and $R$ averaged over all agents, and NLL on the toy dataset. $S  \& R$ is a joint accuracy, where the joint $S  \& R$ is deemed correct if both S and R are correctly inferred for every agent. The first column lists the number of sources $|S|$, and the second row lists the number of agents $|A|$. }
\label{tab:res_toy}
\end{center}
\end{table}

\begin{table}[t!]\footnotesize
\begin{center}
\begin{tabular}{|l|c|c|c|}
\hline
Dataset &  $|S|$ & Source Name \\
\hline
\textcircled{1} Courtyard  &  19 & bench/chair,food truck, bldg, \\
& & vending machine,  trash can, exit \\ \hline
\textcircled{2} SQ1   & 15 & bench/chair, trash can, bldg, exit \\ \hline
\textcircled{3} SQ2   & 22 & bench/chair, trash can, bldg, exit  \\ \hline
\textcircled{4} VIRAT   & 17 & vehicle, exit  \\ \hline
\textcircled{5} CourtyardNew & 16 & bench/chair, exit, bldg \\ \hline
\textcircled{6} AckermanUnion1 & 16 & table, trash can, bldg, exit \\ \hline
\textcircled{7} AckermanUnion2 & 16 & bench/chair, bldg, exit \\ \hline
\textcircled{8} AerialVideo & 16 & table, vehicle, trash can, bldg, exit \\ \hline
\end{tabular}
\end{center}
\vspace{-5pt}
\caption{Summary of functional objects for the datasets}
\label{tab:dataset_source}
\end{table}

\begin{figure*}[t]
\begin{center}
\includegraphics[width=0.8\linewidth]{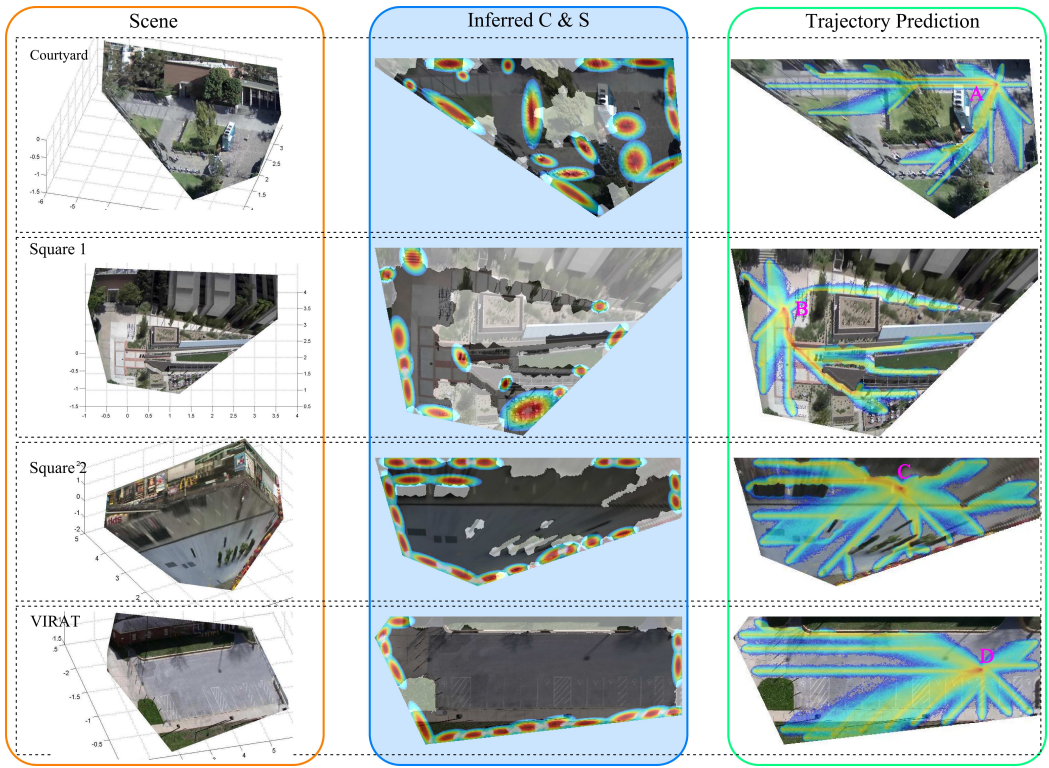}
\end{center}
\vspace{-10pt}
   \caption{Qualitative experiment results for latent functional objects localization and ``single intent'' prediction in 4 scenes. Each row is one scene. The 1st column is the reconstructed 3D surfaces of each scene. The 2nd column is the estimated layout of obstacles (the white masks) and dark matter (the Gaussians). The 3rd column is an example of trajectory prediction by sampling, we predict the future trajectory for a particular agent at some position ($A$, $B$, $C$, $D$) in the scene toward each potential source in $S$, the warm and cold color represent high and low probability of visiting that position respectively.}
\label{fig:res_datasets}
\end{figure*}

\subsection{Real Scenes}\label{sec:real_scenes}
\subsubsection{Datasets}

We use 8 different real scenes for the experiments: \textcircled{1} Courtyard dataset \cite{Amer2012}; video sequences of two squares \textcircled{2} SQ1 and \textcircled{3} SQ2 annotated by VATIC \cite{Vondrick2012}; \textcircled{4} VIRAT ground dataset \cite{Oh2011}; new scenes including CourtyardNew \textcircled{5}, AckermanUnion1 \textcircled{6} and AckermanUnion2 \textcircled{7}; AerialVideo \textcircled{8} from UCLA Aerial Event dataset \cite{Shu2015}. SQ1 is 20min, $800 \times 450$, 15 fps. SQ2 is 20min, $2016 \times 1532$, 12 fps. We use the same scene A of VIRAT as in \cite{Kitani2012}. New videos \textcircled{5}\textcircled{6}\textcircled{7} last 2min, 30min and 30min respectively. We select video 59 from \cite{Shu2015}. The last four videos all have 15 fps and $1920\times 1080$ resolution. For ``single intent'' prediction, we allow the initial observation of 50\% of the video footage, which for example gives about 300 trajectories in \textcircled{1}.

While the ground-truth annotation of ``single'' and ``sequential'' intent behaviors is straightforward, in real scenes, we have encountered a few ambiguous cases of ``change of intent" where different annotators have disagreed about ``single'' or ``change of intent" behavior for the same trajectory. In such ambiguous cases we used a majority vote as ground truth for intent behavior.

\begin{table}\scriptsize
\begin{center}
    \setlength\tabcolsep{2pt}%
\begin{tabular}{|c|cc|cc|ccc|cccc|}
\hline
\multirow{2}{*}{Dataset} & \multicolumn{2}{|c|}{S} & \multicolumn{2}{|c|}{R} & \multicolumn{3}{|c|}{NLL} & \multicolumn{4}{|c|}{MHD}  \\
\cline{2-12} 
& Our & Initial & Our & GM & Our & \cite{Kitani2012} & RW & Our & RW & SP & PM \\ \hline
\textcircled{1} & {\bf 0.89} & 0.23 & {\bf 0.52} & 0.31 & {\bf 1.635} & - & 2.197 & {\bf 14.8} & 243.1 & 43.2 & 207.5  \\ \hline
\textcircled{2} & {\bf 0.87} & 0.37 & {\bf 0.65} & 0.53 & {\bf 1.459} & - & 2.197 & {\bf 11.6} & 262.1 & 39.4 & 237.9  \\ \hline
\textcircled{3} & {\bf 0.93} & 0.26 & {\bf 0.49} & 0.42 & {\bf 1.621} & - & 2.197 & {\bf 21.5} & 193.8 & 27.9 & 154.2  \\ \hline
\textcircled{4} & {\bf 0.95} & 0.25 & {\bf 0.57} & 0.46 & {\bf 1.476} & 1.594 & 2.197 & {\bf 16.7} & 165.4 & 21.6 & 122.3  \\ \hline
\textcircled{5} & {\bf 0.81} & 0.19 & {\bf 0.75} & 0.50 & {\bf 0.243} & - & 2.197 & {\bf 20.1} & 171.3 & 27.5 & 181.7  \\ \hline
\textcircled{6} & {\bf 0.81} & 0.38 & {\bf 0.63} & 0.42 & {\bf 0.776} & - & 2.197 & {\bf 13.7} & 183.0 & 17.5 & 128.6  \\ \hline
\textcircled{7} & {\bf 0.88} & 0.32 & {\bf 0.56} & 0.39 & {\bf 1.456} & - & 2.197 & {\bf 19.3} & 150.1 & 26.9 & 119.2  \\ \hline
\textcircled{8} & {\bf 0.63} & 0.25 & {\bf 0.67} & 0.33 & {\bf 1.710} & - & 2.197 & {\bf 15.9} & 289.3 & 26.5 & 137.0  \\ \hline
\end{tabular}
\caption{Qualitative results of ``single intent''}
\vspace{-10pt}
\label{tab:res_real}
\end{center}
\end{table}
\begin{table}
\centering
\begin{tabular}{|c|c|c|c|c|}
\hline
Obs. $\%$  & S & R& NLL & MHD  \\ \hline
$45\%$ & 0.85 & 0.47 & 1.682 & 15.7 \\ \hline
$40\%$ & 0.79 & 0.41 & 1.753 & 16.2 \\ \hline

\end{tabular}
\caption{Results on \textcircled{1} with different observed ratios}
\label{tab:obs_ratio}
\end{table}


\begin{figure*}
\centering
\begin{minipage}{0.73\linewidth}
\begin{center}
\captionsetup{width=0.9\textwidth}
\includegraphics[width=1.0\linewidth]{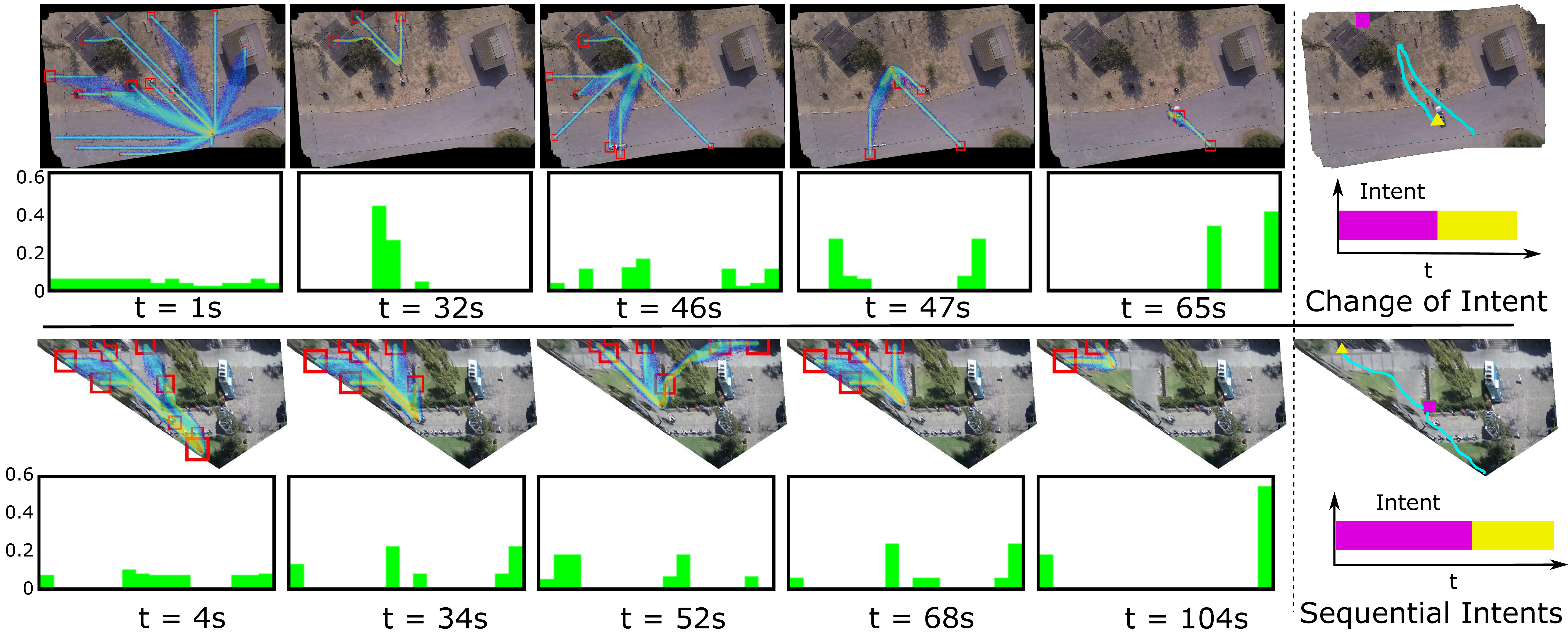}
\end{center}
\vspace{-5pt}
   \caption{Qualitative result of ``sequential intents'' and ``change of intent''. Left: online prediction where the red bounding boxes represent the possible intents at a certain moment and a larger bounding box indicates an intent with higher probability; histograms represent the probabilities of each functional object being the intent of the agent at a given moment. Right: offline intent types and intents inference based on the full observation of trajectories, where the square is the first intent and the triangle is the second intent.}
\label{fig:aerial_video}
\end{minipage}
\begin{minipage}{0.25\linewidth}
\captionsetup{width=0.95\textwidth}
\centering
\begin{subfigure}[b]{0.95\linewidth}
\captionsetup{width=1\textwidth}
\includegraphics[trim={30, 225, 60, 245}, clip, width=1.0\linewidth]{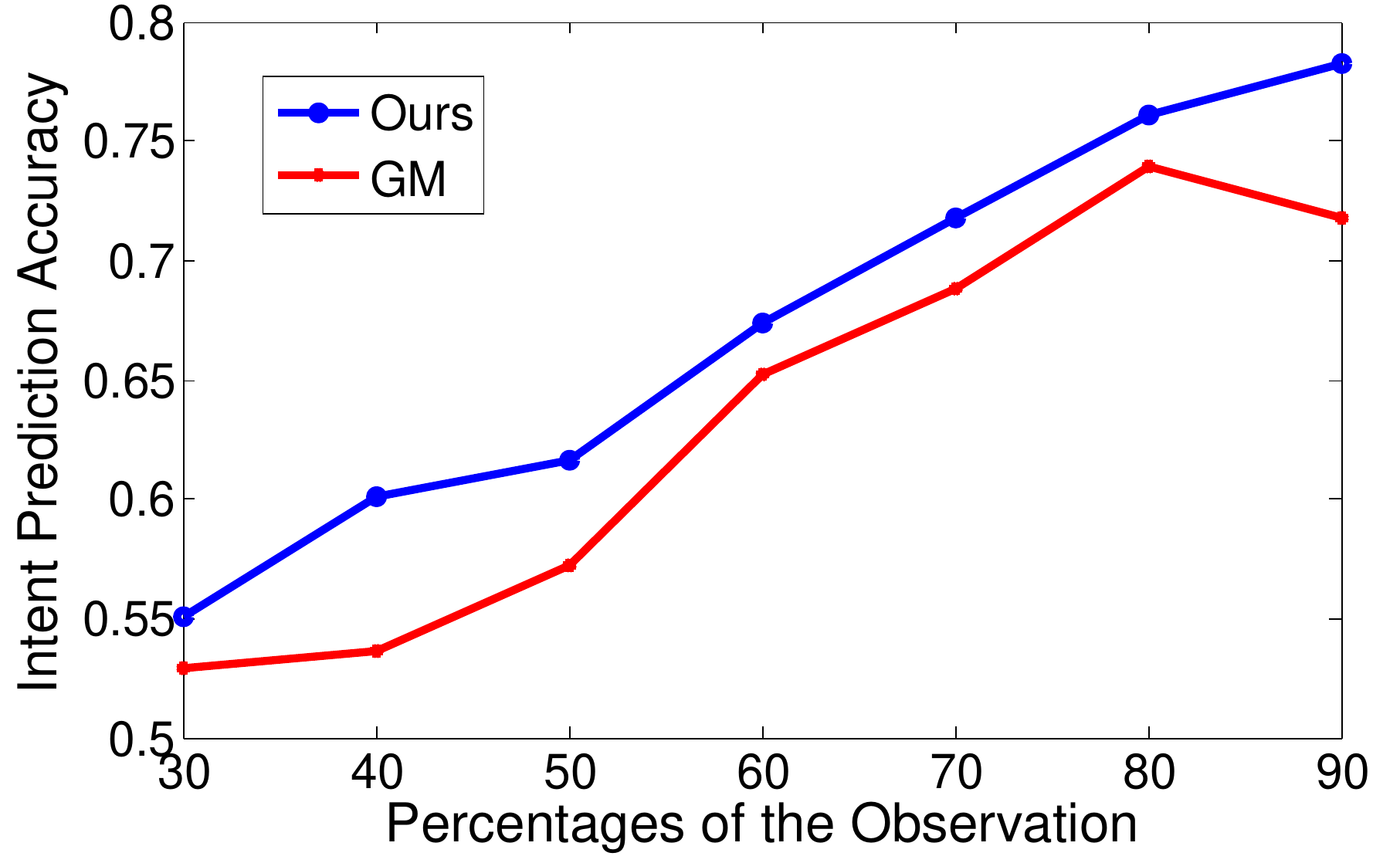}
\vspace{-15pt}
   \caption{Online intent prediction accuracy.}
\label{fig:online_intent_prediction}
\end{subfigure}
\begin{subfigure}[b]{0.95\linewidth}
\captionsetup{width=1\textwidth}
\includegraphics[trim={40, 230, 60, 245}, clip, width=1.0\linewidth]{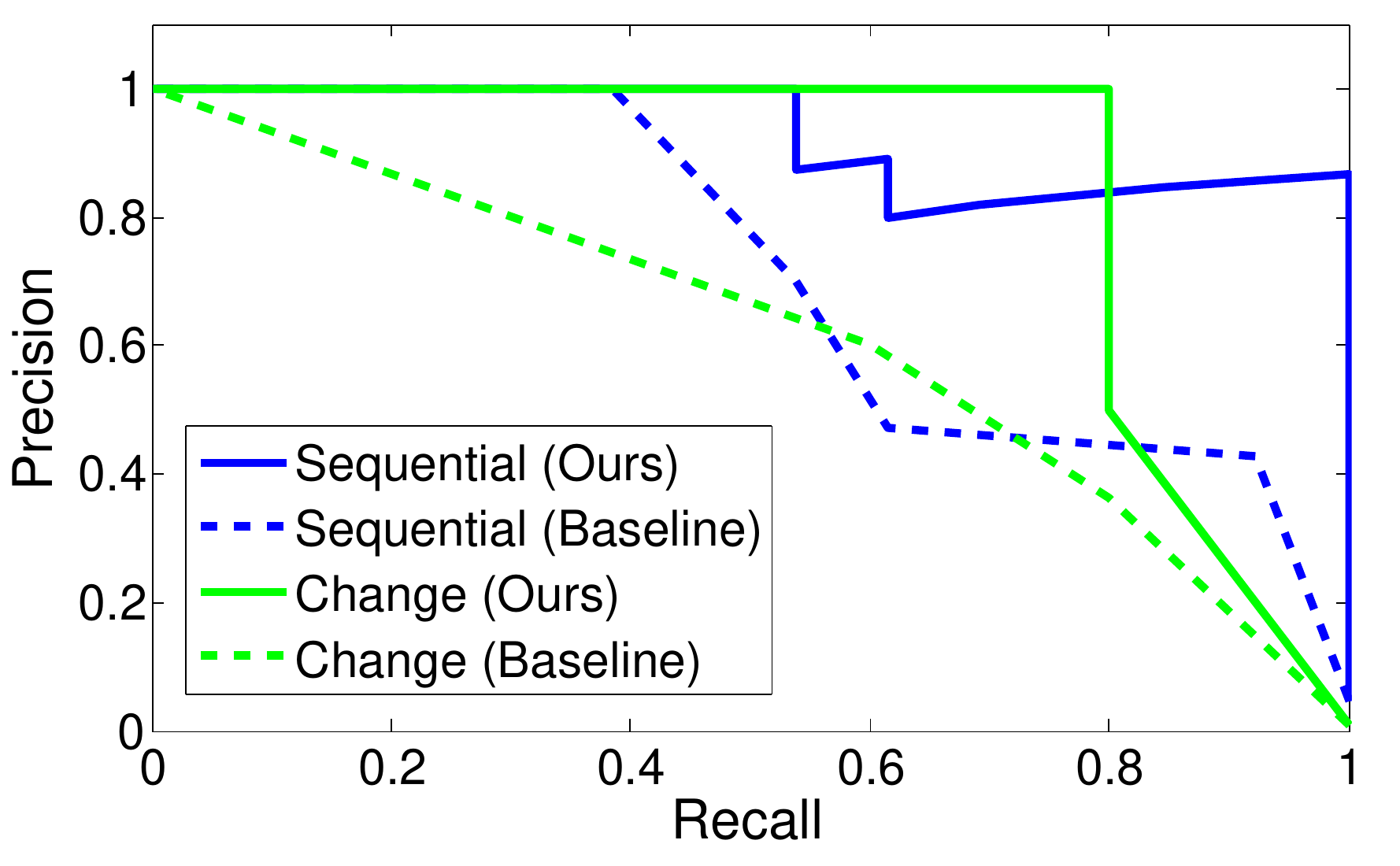}
\vspace{-15pt}
   \caption{Offline intent type inference.}
\label{fig:offline_intent_type}
\end{subfigure}
\caption{PR curves of ``sequential intents'' and ``change of intent''.}
\label{fig:complex_change}
\vspace{-5pt}
\end{minipage}
\end{figure*} 

\begin{figure*}[t]
\centering
\begin{subfigure}[b]{0.3\textwidth}
\includegraphics[width = \linewidth]{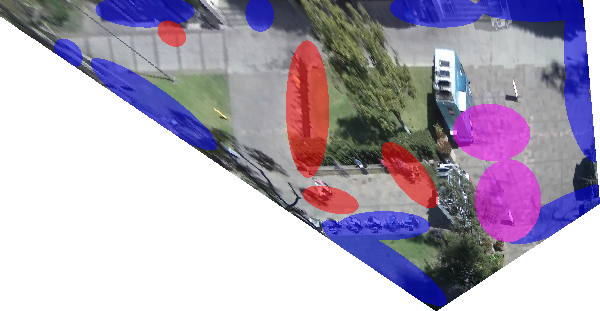}
\vspace{-15pt}
\caption{Courtyard}
\label{fig:func_map_1}
\end{subfigure}
~
\begin{subfigure}[b]{0.2\textwidth}
\includegraphics[width = \linewidth]{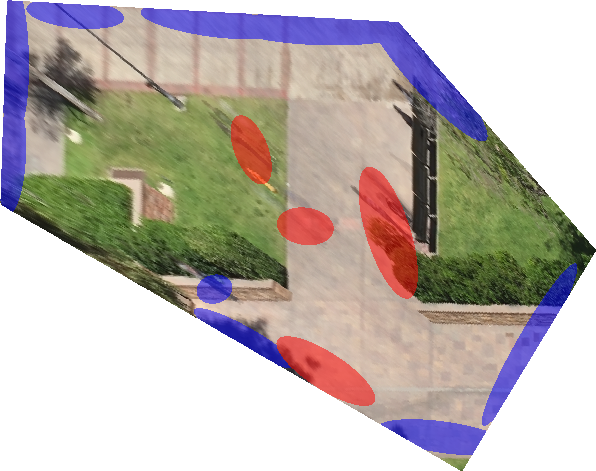}
\vspace{-15pt}
\caption{CourtyardNew}
\label{fig:func_map_2}
\end{subfigure}
~
\begin{subfigure}[b]{0.19\textwidth}
\includegraphics[width = \linewidth]{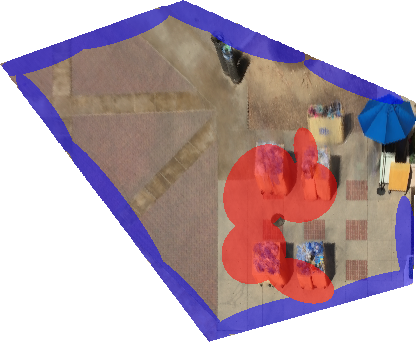}
\vspace{-15pt}
\caption{AckermanUnion1}
\label{fig:func_map_3}
\end{subfigure}
~
\begin{subfigure}[b]{0.2\textwidth}
\includegraphics[width = \linewidth]{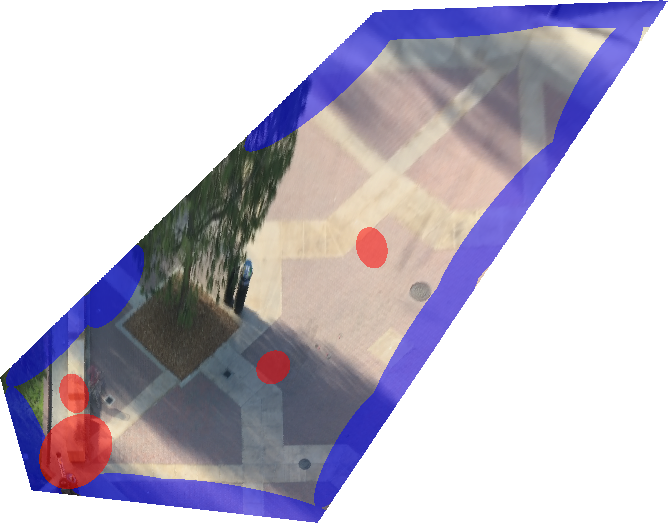}
\vspace{-15pt}
\caption{AckermanUnion2}
\label{fig:func_map_4}
\end{subfigure}
\caption{Functional object clustering results. Each ellipse map is an inferred functional object and the four different colors (i.e., magenta, red, blue, and yellow) represent four latent functional classes.}
\label{fig:func_map}
\end{figure*}

\subsubsection{Experiment 1: Functional Object Localization and Single Intent Prediction}
In this experiment, we use only ``single intent'' trajectories of real-world datasets. Our qualitative results are shown in Figure \ref{fig:res_datasets}, and quantitative evaluation is presented in Tab. \ref{tab:res_real}. As can be seen: (1) We are relatively insensitive to the specific choice of model parameters. (2) We handle challenging scenes with arbitrary layouts of dark matter, both in the middle of the scene and at its boundaries. From Tab. \ref{tab:res_real}, the comparison with the baselines demonstrates that the initial guess of sources based on partial observations gives very noisy results. These noisy results are significantly improved in our DD-MCMC inference. Also, our method is a slightly better than the baseline GM if there are a few obstacles in the middle of the scene. But we get a huge performance improvement over GM if there are complicated obstacles in the scene. This shows that our global plan based relation prediction is better than GM. 

Based on $S$ and $C$, we model human motion probabilistically given their goals and understanding of scenes so that we can predict their future trajectories with probability. The prediction accuracy on the four scenes are summarized in Table~\ref{tab:res_real}. It appears that our results are also superior to the random walk. The baselines RW and PM produce bad trajectory prediction. While SP yields good results for scenes with a few obstacles, it is brittle for more complex scenes which we successfully handle. When the size of $S$ is large (e.g., many exists from the scene), our estimation of human goals may not be exactly correct. However, in all these error cases, the goal that we estimate is not spatially far away from the true goal. Also, in these cases, the predicted trajectories are also not far away from the true trajectories measured by MHD and NLL. Our performance downgrades gracefully with the reduced observation time as Table~\ref{tab:obs_ratio} indicates. We outperform the state of the art \cite{Kitani2012}. Note that the MHD absolute values produced by our approach and \cite{Kitani2012} are not comparable, because this metric is pixel based and depends on the resolution of reconstructed 3D surface.  

Our results show that our method successfully addresses surveillance scenes of various complexities. 
 
\subsubsection{Experiment 2: ``Sequential Intents'' and ``Change of Intent'' Inference}

For the evaluation, we randomly select 500 trajectories from all scenes and manually annotate the types of intent for each of them. In all these trajectories, there are 13 ``sequential intents'' instances, 5 ``change of intent'' instances, and the remaining trajectories all have ``single intent''.

The qualitative results of both online prediction and offline inference are visualized in Fig.~\ref{fig:aerial_video}. Note that we automatically infer the vehicles as functional objects in \textcircled{8}. For the ``change of intent'' case, the agent switched the intent from a building to a vehicle at around 46s; for the ``sequential intents'' case, the agent first went a trash can near the bench (latent behind the bush) to throw trash (52s), and then left for one of the exits (68s). It appears that i) our online predictions can reflect the intent changes of the observed agents and ii) with the full trajectory observation, the offline inference correctly recognizes the intent types and the corresponding temporal passing of the agents' intents.

The online intent prediction accuracy is shown in Fig.~\ref{fig:online_intent_prediction}, which indicates that ours consistently outperform the baseline. In Fig.~\ref{fig:offline_intent_type}, we plot the precision-recall curves for the offline intent type inference. The baseline method uses a motion-based measurement of the trajectory, i.e., the longest stationary time before arriving the intent for the ``sequential intents'', and the largest turning angle for the ``change of intent''. Ours yields much higher mAPs for both ``sequential intents'' and ``change of intent'' (0.93 and 0.80 respectively) than the baseline (0.66 and 0.43 respectively).



\begin{figure}[t]
\begin{center}
\includegraphics[trim={40, 10, 50, 0},clip,width=0.80\linewidth]{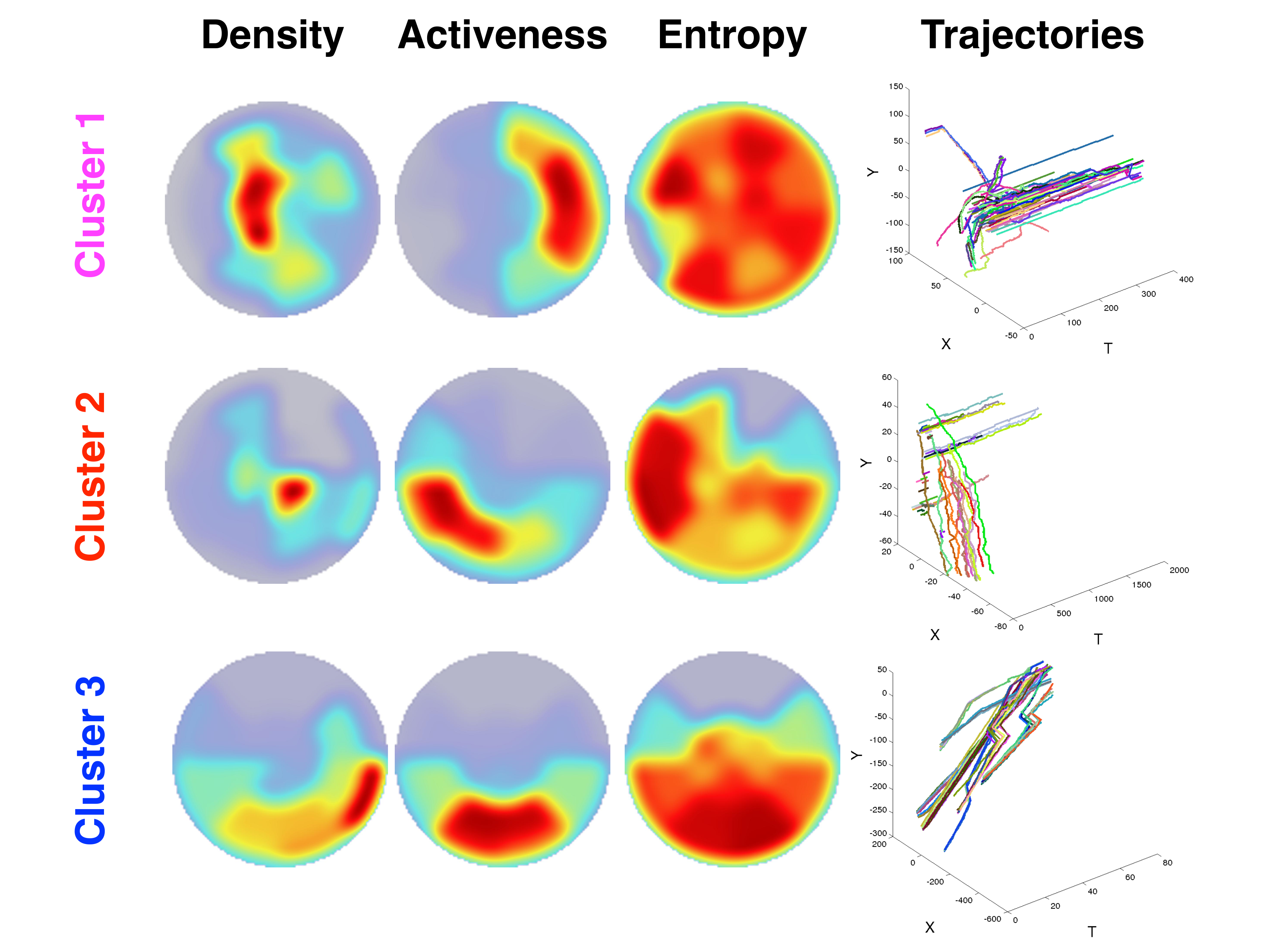}
\end{center}
\vspace{-10pt}
   \caption{Visualization of mean feature maps for the three clusters of functional objects and the associated trajectories of an example functional object within each cluster. Note that all the functional objects within the same cluster are aligned to a reference direction.}
\label{fig:feat_maps}
\end{figure}

\subsubsection{Experiment 3: Functional Objects Clustering}
The estimated trajectories of people can be viewed as a summary of their behavior under the attraction and repulsion forces of functional objects in the scene. We show that it is possible to cluster functional objects (of different semantic classes) by the properties of human behavior, and in this way {\em discover} functional classes. 

For such clustering we use the following three features characterizing the nearby region of each identified ``dark matter" in a scene: a) the density of the associated agents around the latent functional objects, i.e., a density map; b) the spatial distribution of the average velocity magnitude of the associated agents, i..e, an activeness map; c) the spatial distribution of the entropy of moving directions of the associated agents, i.e., an entropy map. Here, we assume that all agents have one single intent and associate their trajectories with their own intents for computing the features. Similar to the shape context features \cite{belongie2002}, we can convert these three maps into three histograms and concatenate them into a feature vector. Based on the feature vectors, we then perform the K-means clustering to group the inferred functional objects. Note that feature maps are aligned within the same cluster by rotation and mirroring. 

We cluster all the functional objects in the real scenes into 3 latent functional classes. Fig.~\ref{fig:func_map} shows the clustering result in 4 scenes and Fig.~\ref{fig:feat_maps} visualizes the mean feature maps of each cluster and the associated trajectories around a example functional object in each cluster. The visualized results confirm that without the appearance and geometry information of the functional objects, we are able to discover meaningful functional classes by analyzing human behaviors, which show clear semantic meanings:  i) magenta regions: queuing areas; ii) red regions: areas where people stand or sit for a long time (e.g., benches, chairs, tables, vending machines); iii) blue regions: exits or buildings. Note that sometimes there are a few small objects in the scenes, e.g., trash cans in Fig.~\ref{fig:func_map_3}, that are not identified as functional objects by our approach, simply because they were never used by the agents in the videos. Interestingly, we occasionally obtain one to two red regions around the lawn (e.g. Fig.~\ref{fig:func_map_2}) or on the square (e.g. Fig.~\ref{fig:func_map_4}) since multiple agents were standing there for a long time in the videos. 

\section{Conclusion}\label{sec:conclusion}
We have addressed the problem of predicting human trajectories in unobserved parts of videos of public spaces, given access only to an initial excerpt of the videos in which most of the human trajectories have not yet reached their respective destinations. We have formulated this problem as that of reasoning about latent human intents to approach functional objects in the scene, and, consequently, identifying a functional map of the scene. Our work extends the classical Lagrangian mechanics to model the scene as a physical system wherein: i) functional objects exert attraction forces on people's motions, and ii) people are viewed as agents who can have intents to approach particular functional objects. For evaluation we have used the benchmark VIRAT, UCLA Courtyard and UCLA Aerial Event datasets, as well as our five videos of public spaces. We have shown that  it is possible to cluster functional objects of different semantic classes by the properties of human motion behavior in their surrounding, and in this way discover functional classes. One limitation of our method is that it does not account for social interactions between agents, which seems a promising direction for future work.


%

%

%

\ifCLASSOPTIONcaptionsoff
  \newpage
\fi

\bibliographystyle{IEEEtran}
\bibliography{pami2016_darkmatter}
\vspace{-40pt}
\begin{IEEEbiography}[{\includegraphics[width=1in,height=1.25in,clip,keepaspectratio]{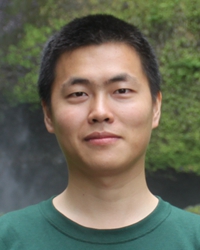}}]{Dan Xie} received the PhD degree in Statistics from University of California, Los Angeles (UCLA) in 2016. He received his B.E. degree in Software Engineering from Zhejiang University in China in 2011. His research interests include computer vision and machine learning.
\end{IEEEbiography}
\vspace{-40pt}
\begin{IEEEbiography}[{\includegraphics[width=1in,height=1.25in,clip,keepaspectratio]{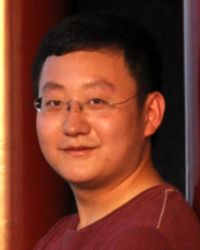}}]{Tianmin Shu}
received his B.S. degree in electronic engineering from Fudan University, China in 2014. He is currently a Ph.D. candidate in the Department of Statistics at University of California, Los Angeles. His research interests include computer vision, robotics and computational cognitive science, especially social scene understanding and human-robot interactions.
\end{IEEEbiography}
\vspace{-40pt}
\begin{IEEEbiography}[{\includegraphics[width=1in,height=1.25in,clip,keepaspectratio]{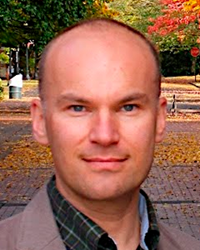}}]{Sinisa Todorovic} received the PhD degree in electrical and computer engineering from the University of Florida in 2005. He was a postdoctoral researcher in the Beckman Institute at the University of Illinois at Urbana-Champaign, between 2005 and 2008. Currently, he is an associate professor in the School of Electrical Engineering and Computer Science at Oregon State University. His research interests include computer vision and machine learning problems.
\end{IEEEbiography}
\vspace{-40pt}
\begin{IEEEbiography}[{\includegraphics[width=1in,height=1.25in,clip,keepaspectratio]{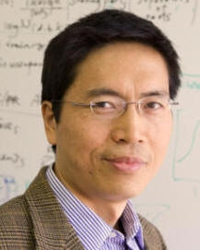}}]{Song-Chun Zhu}
received his Ph.D. degree
from Harvard University. He is currently professor of Statistics and Computer Science at UCLA. His research interests include computer vision, statistical modeling, learning, cognition, and robot autonomy. He received a number of honors, including the Helmholtz Test-of-time award in ICCV 2013, the Aggarwal prize from the IAPR in 2008, the David Marr Prize in 2003 with Z. Tu et al. for image parsing, twice Marr Prize honorary nominations with Y. Wu et al. in 1999 for texture modeling and 2007 for object modeling respectively. He received the Sloan Fellowship in 2001, a US NSF Career Award in 2001, and an US ONR Young Investigator Award in 2001. He is a Fellow of IEEE since 2011.
\end{IEEEbiography}

\end{document}